# A Survey of Convolutional Neural Networks: Analysis, Applications, and Prospects

Zewen Li, Wenjie Yang, Shouheng Peng, Fan Liu, *Member, IEEE*

***Abstract*—Convolutional Neural Network (CNN) is one of the most significant networks in the deep learning field. Since CNN made impressive achievements in many areas, including but not limited to computer vision and natural language processing, it attracted much attention both of industry and academia in the past few years. The existing reviews mainly focus on the applications of CNN in different scenarios without considering CNN from a general perspective, and some novel ideas proposed recently are not covered. In this review, we aim to provide novel ideas and prospects in this fast-growing field as much as possible. Besides, not only two-dimensional convolution but also one-dimensional and multi-dimensional ones are involved. First, this review starts with a brief introduction to the history of CNN. Second, we provide an overview of CNN. Third, classic and advanced CNN models are introduced, especially those key points making them reach state-of-the-art results. Fourth, through experimental analysis, we draw some conclusions and provide several rules of thumb for function selection. Fifth, the applications of one-dimensional, two-dimensional, and multi-dimensional convolution are covered. Finally, some open issues and promising directions for CNN are discussed to serve as guidelines for future work.**

***Index Terms*—Deep learning, convolutional neural networks, deep neural networks, computer vision.**

## I. Introduction

CONVOLUTIONAL Neural Network (CNN) has been making brilliant achievements. It has become one of the most representative neural networks in the field of deep learning. Computer vision based on convolutional neural networks has enabled people to accomplish what had been considered impossible in the past few centuries, such as face recognition, autonomous vehicles, self-service supermarket, and intelligent medical treatment. To better understand modern convolutional neural network and make it better serve human beings, in this paper, we present an overview of CNN, introduce classic models and applications, and propose some prospects for CNN.

The emergence of convolutional neural networks cannot be separated from Artificial Neural Networks (ANN). In 1943, McCulloch and Pitts [1] proposed the first mathematical model of neurons—the MP model. In the late 1950s and early 1960s, Rosenblatt [2], [3] proposed a single-layer perceptron model by adding learning ability to the MP model. However, single-layer perceptron network cannot handle linear inseparable problems (such as XOR problems). In 1986, Hinton *et al.* [4] proposed a multi-layer feedforward network trained by the error back-propagation algorithm—Back Propagation Network (BP Network), which addressed some problems that single-layer perceptron could not solve. In 1987, Waibel *et al.* [5] proposed Time Delay Neural Network (TDNN) for speech recognition, which can be viewed as a one-dimensional convolutional neural network. Then, Zhang [6] proposed the first two-dimensional convolutional neural network—Shift-invariant Artificial Neural Network (SIANN). LeCun *et al.* [7] also constructed a convolutional neural network for handwritten zip code recognition in 1989 and used the term "convolution" firstly, which is the original version of LeNet. In the 1990s, various shallow neural networks were successively proposed, such as Chaotic neural networks [8] and A general regression neural network [9]. The most famous one is LeNet-5 [10]. Nevertheless, when the number of layers of neural networks is increased, traditional BP networks would encounter local optimum, overfitting, gradient vanishing, and gradient exploding problems. In 2006, Hinton *et al.* [11] proposed the following points: 1) Multi-hidden layers artificial neural networks have excellent feature learning ability; 2) The "layer-wise pre-training" can effectively overcome the difficulties of training deep neural networks, which brought about the study of deep learning. In 2012, Alex *et al.* [11] achieved the best classification result at that time using deep CNN in the ImageNet Large Scale Visual Recognition Challenge (LSVRC), which attracted researchers much of attention and greatly promoted the development of modern CNN.

Before our work, there exist several researchers reviewed CNN. Aloysius *et al.* [12] paid attention to frameworks of deep learning chronologically. Nevertheless, they did not fully explain why these architectures are better than their predecessors and how these architectures achieved their goals. Dhillon *et al.* [13] discussed the architectures of some classic networks, but there are many new-generation networks, after their work, have been proposed, such as MobileNet v3, Inception v4, and ShuffleNet series, which deserve researchers' attention. Besides, the work reviewed applications of CNN for object detection. Rawat *et al.* [14] reviewed CNN for image recognition. Liu *et al.* [15] discussed CNN for image

This work was supported in part by National Natural Science Foundation of China under grant No. 61602150, Natural Science Foundation of Jiangsu Province under grant No. BK20191298. (Corresponding author: Fan Liu)

Zewen Li, Wenjie Yang, Shouheng Peng, and Fan Liu are with College of Computer and Information, Hohai University, Nanjing, 210098, China (servon@hhu.edu.cn, vicent@hhu.edu.cn, shoehengpeng@hhu.edu.cn, fanliu@hhu.edu.cn)

recognition. Ajmal *et al.* [16] discussed CNN for image segmentation. These reviews mentioned above mainly reviewed the applications of CNN in different scenarios without considering CNN from a general perspective. Also, due to the rapid development of CNN, lots of inspiring ideas in this field have been proposed, but these reviews did not fully cover them.

In this paper, we focus on analyzing and discussing CNN. In detail, the key contributions of this review are as follows: 1) We provide a brief overview of CNN, including some basic building blocks of modern CNN, in which some fascinating convolution structures and innovations are involved. 2) Some classic CNN-based models are covered, from LeNet-5, AlexNet to MobileNet v3 and GhostNet. Innovations of these models are emphasized to help readers draw some useful experience from masterpieces. 3) Several representative activation functions, loss functions, and optimizers are discussed. We reach some conclusions about them through experiments. 4) Although applications of two-dimensional convolution are widely used, one-dimensional and multi-dimensional ones should not be ignored. Some of typical applications are presented. 5) We raise several points of view on prospects for CNN. Part of them are intended to refine existing CNNs, and the others create new networks from scratch.

We organize the rest of this paper as follows: Section 2 takes an overview of modern CNN. Section 3 introduces many representative and classic CNN-based models. We mainly focus on the innovations of these models, but not all details. Section 4 discusses some representative activation functions, loss functions, and optimizers, which can help readers select them appropriately. Section 5 covers some applications of CNN from the perspective of different dimensional convolutions. Section 6 discusses current challenges and several promising directions or trends of CNN for future work. Section 7 concludes the survey by giving a bird view of our contributions.

## II. BRIEF OVERVIEW OF CNN

Convolutional neural network is a kind of feedforward neural network that is able to extract features from data with convolution structures. Different from the traditional feature extraction methods [17], [18], [19] , CNN does not need to extract features manually. The architecture of CNN is inspired by visual perception [20]. A biological neuron corresponds to an artificial neuron; CNN kernels represent different receptors that can respond to various features; activation functions simulate the function that only neural electric signals exceeding a certain threshold can be transmitted to the next neuron. Loss functions and optimizers are something people invented to teach the whole CNN system to learn what we expected. Compared with general artificial neural networks, CNN possesses many advantages: 1) Local connections. Each neuron is no longer connected to all neurons of the previous layer, but only to a small number of neurons, which is effective in reducing parameters and speed up convergence; 2) Weight sharing. A group of connections can share the same weights, which reduces parameters further. 3) Down-sampling dimensionality reduction. A pooling layer harnesses the principle of image local correlation to down-sample an image, which can reduce the amount of data while retaining useful information. It can also reduce the number of parameters by removing trivial features. The three appealing characteristics make CNN become one of the most representative algorithms in the deep learning field.

To be specific, in order to build a CNN model, four components are typically needed. Convolution is a pivotal step for feature extraction. The outputs of convolution can be called feature maps. When setting a convolution kernel with a certain size, we will lose information in the border. Hence, padding is introduced to enlarge the input with zero value, which can adjust the size indirectly. Besides, for the sake of controlling the density of convolving, stride is employed. The larger the stride, the lower the density. After convolution, feature maps consist of a large number of features that is prone to causing overfitting problem [21]. As a result, pooling [22] (a.k.a. down-sampling) is proposed to obviate redundancy, including max pooling and average pooling. The procedure of a CNN is shown in Fig. 1.

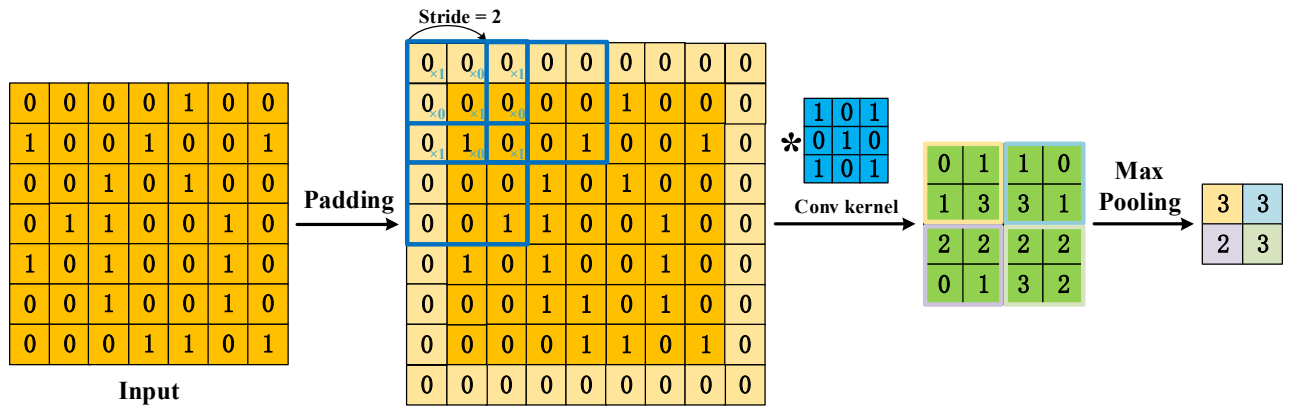

Fig. 1. Procedure of a two-dimensional CNN

Furthermore, in order for convolution kernels to perceive larger area, dilated convolution [23] was proposed. A general 3 × 3 convolution kernel is shown in Fig. 2 (a), and a 2-dilated 3 × 3 convolution kernel and a 4-dilated 3 × 3 convolution kernel are shown in Fig. 2 (b) and (c). Note that there is an empty value (filling with zero) between each convolution kernel point. Even though the valid kernel points are still 3 × 3, a 2-dilated convolution has a 7 × 7 receptive field, and a 4-dilated convolution has a 15 × 15 receptive field.

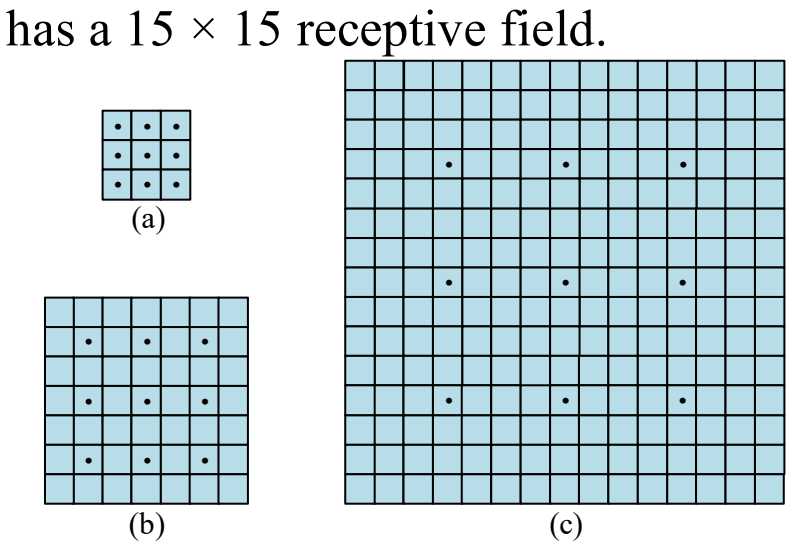

Fig. 2. Comparison between general convolution kernel and dilated convolution kernel. (a) A general 3 × 3 convolution kernel (b) A 2-dilated 3 × 3 convolution kernel (c) A 4-dilated 3 × 3 convolution kernel

As shown in Fig. 3, deformable convolution [23] was proposed to handle the problem that the shape of objects in the real world are usually irregular. Deformable convolution is able to only focus on what they are interested in, making the feature maps are more representative.

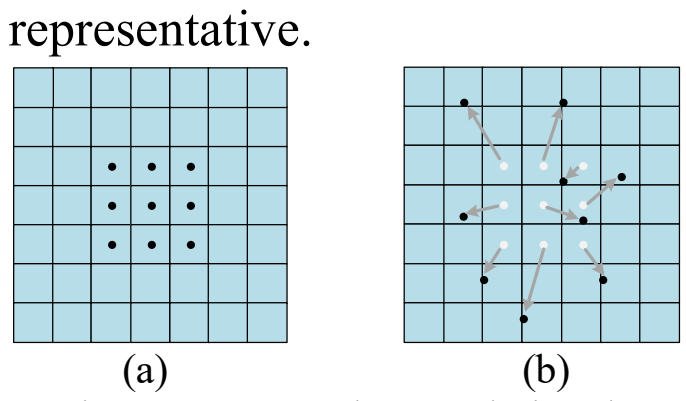

Fig. 3. Comparison between general convolution kernel and deformable convolution kernel. (a) A general 3 × 3 convolution kernel (b) A deformable 3 × 3 convolution kernel

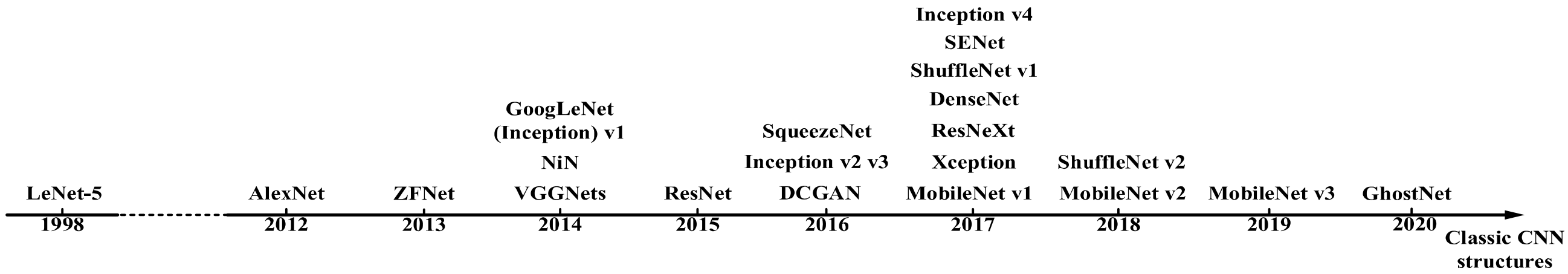

Fig. 4. Part of classic CNN models. NiN: Network in Network; ResNet: Residual Netwrok; DCGAN: Deep Convolutional Generative Adversarial Network; SENet: Squeeze-and-Excitation Network

Moreover, there exist a variety of awesome convolutions, such as Separable convolutions [24], [25], [26], [27], [28], group convolutions [11], [29], [30], [31] and multi-dimensional convolutions, which are discussed in Section 3 and Section 5.

## III. Classic CNN Models

Since AlexNet was proposed in 2012, researchers have invented a variety of CNN models—deeper, wider, and lighter. Part of well-known models can be seen in Fig. 4. Due to the limitation of paper length, this section aims to take an overview of several representative models, and we will emphatically discuss the innovations of them to help readers understand the main points and propose their own promising ideas.

### A. LeNet-5

LeCun *et al.* [10] proposed LeNet-5 in 1998, which is an efficient convolutional neural network trained with the backpropagation algorithm for handwritten character recognition. As shown in Fig. 5, LeNet-5 is composed of seven trainable layers containing two convolutional layers, two pooling layers, and three fully-connected layers. LeNet-5 is the pioneering convolutional neural network combining local receptive fields, shared weights, and spatial or temporal sub-sampling, which can ensure shift, scale, and distortion invariance to some extent. It is the foundation of modern CNN. Although LeNet-5 is useful for recognizing handwriting characters and reading bank checks, it still does not exceed the traditional support vector machine (SVM) and boosting algorithms. As a result, LeNet-5 did not obtain enough attention at that time.

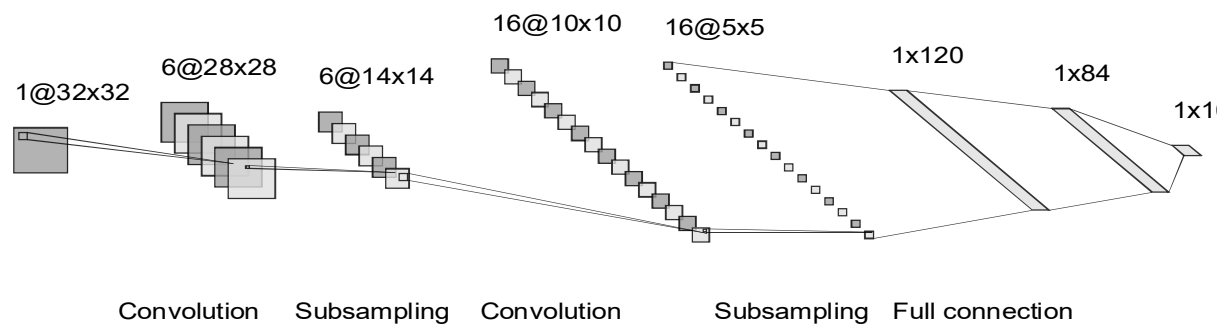

Fig. 5. Architecture of LeNet-5

### B. AlexNet

Alex *et al.* [11] proposed the AlexNet in 2012, which won the championship in the ImageNet 2012 competition. As shown in Fig. 6, AlexNet has eight layers, containing five convolutional layers and three fully-connected layers.

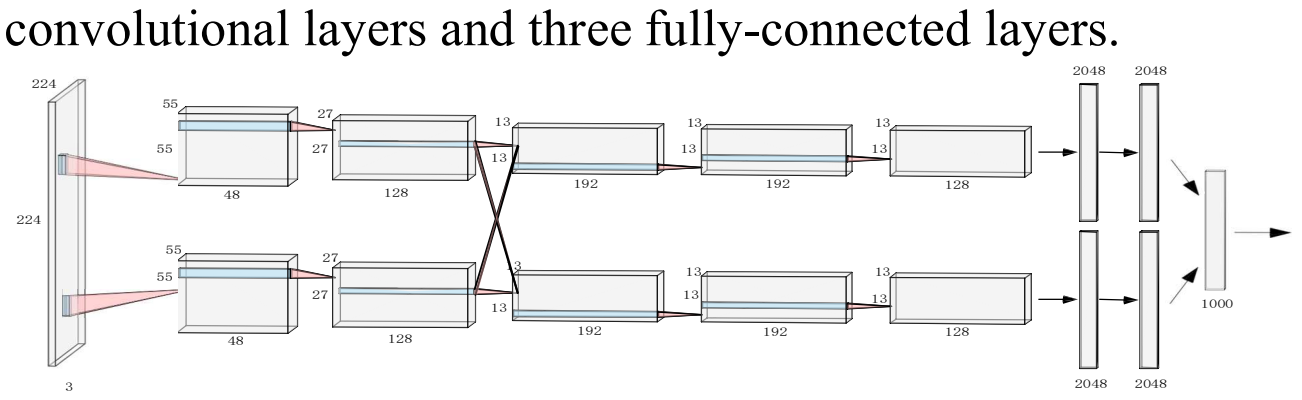

Fig. 6. Architecture of AlexNet

AlexNet carries forward LeNet's ideas and applies the basic principles of CNN to a deep and wide network. It successfully leverages ReLU activation function, dropout, and local response normalization (LRN) for the first time on CNN. At the same time, AlexNet also makes use of GPUs for computing acceleration. The main innovations of AlexNet lie in the following:

1) AlexNet uses ReLU as the activation function of CNN, which mitigates the problem of gradient vanishing when the network is deep. Although the ReLU activation function was proposed long before AlexNet, it was not carried forward until the appearance of AlexNet.

2) Dropout is used by AlexNet to randomly ignore some neurons during training to avoid overfitting. This technique is mainly used in the last few fully-connected layers.

3) In convolutional layers of AlexNet, overlapping max pooling is used to replace average pooling that was commonly-used in the previous convolutional neural networks. Max pooling can avoid the blurred result of average pooling, and overlapping pooling can improve the richness of features.

4) LRN is proposed to simulate the lateral inhibition mechanism of the biological nervous system, which means the neuron receiving stimulation can inhibit the activity of peripheral neurons. Similarly, LRN can make neurons with small values are suppressed, and those with large values are relatively active, the function of which is very similar to normalization. Hence, LRN is a way to enhance the generalization ability of the model.

5) AlexNet also employs two powerful GPUs to train group convolutions. Since the computing resource limit of one GPU, AlexNet designs a group convolution structure, which can be trained on two distinct GPUs. And then, two feature maps generated by two GPUs can be combined as the final output.

6) AlexNet adopts two data augmentation methods in training. The first is extracting random $224 \times 224$ patches from the original $256 \times 256$ images and their horizontal reflections to obtain more training data. Besides, the Principal Component Analysis (PCA) is utilized to change the RGB values of the training set. When making predictions, AlexNet also enlarges the dataset and then calculate the average of their predictions as the final result. AlexNet shows that the use of data augmentation can substantially mitigate overfitting problem and improve generalization ability.

### C. VGGNets

VGGNets [32] are a series of convolutional neural network algorithms proposed by the Visual Geometry Group (VGG) of Oxford University, including VGG-11, VGG-11-LRN, VGG-13, VGG-16, and VGG-19. VGGNets secured the first place in the localization track of ImageNet Challenge 2014. The authors of VGGNets prove that increasing the depth of neural networks can improve the final performance of the network to some extent. Compared with AlexNet, VGGNets have the following improvements:

1) The LRN layer was removed since the author of VGGNets found the effect of LRN in deep CNNs was not obvious.

2) VGGNets use 3 × 3 convolution kernels rather than 5 × 5 or 5 × 5 ones, since several small kernels have the same receptive field and more nonlinear variations compared with larger ones. For instance, the receptive field of 3 × 3 kernels is the same as one 5 × 5 kernel. Nevertheless, the number of parameters reduces by about 45%, and three kernels have three nonlinear variations.

### *D. GoogLeNet*

GoogLeNet [33] is the winner of the ILSVRC 2014 image classification algorithms. It is the first large-scale CNN formed by stacking with Inception modules. Inception networks have four versions, namely Inception v1 [33], Inception v2 [34], [35], Inception v3 [35], and Inception v4 [36].

#### *1) Inception v1*

Due to objects in images have different distances to cameras, an object with a large proportion of an image usually prefers a large convolution kernel or a few small ones. However, a small object in an image is the opposite. Based on the past experience, large kernels have many parameters to train, and deep networks are hard to train. As a result, Inception v1 [33] deploys 1 × 1, 3 × 3, 5 × 5 convolution kernels to construct a “wide” network, which can be seen in Fig. 7, Convolution kernels with different sizes can extract the feature maps of different scales of the image. Then, those feature maps are stacked to obtain a more representative one. Besides, 1 × 1 convolution kernel is used to reduce the number of channels, i.e., reduce computational cost.

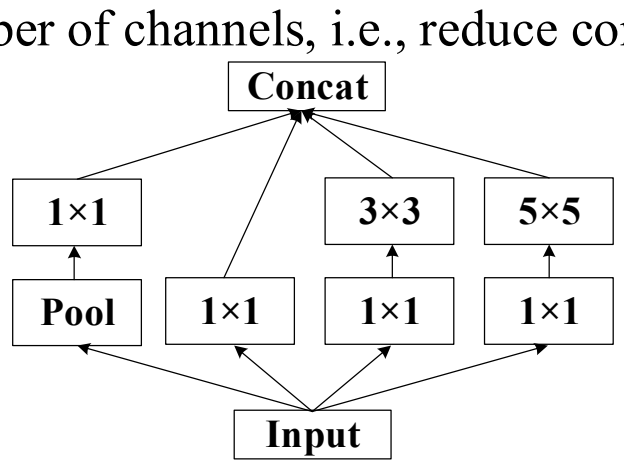

Fig. 7. Inception v1 module with dimension reductions

#### *2) Inception v2*

Inception v2 [35] utilizes batch normalization to handle internal covariate shift problem [34]. The output of every layer is normalized to normal distribution, which can increase the robustness of the model and train the model with a relatively large learning rate.

Furthermore, Inception v2 shows that a single 5 × 5 convolutional layers can be replaced by two 3 × 3 ones, shown in Fig. 8 (a). One n x n convolutional layer can be replaced by one 1 x n and one n x 1 convolutional layer shown in Fig. 8 (b). However, the original paper points out the use of factorization is not effective in the early layers. It is better to use it on medium-sized feature maps. And filter banks should be expanded (wider but not deeper) to improve high dimensional representations. Hence, only the last 3 × 3 convolution of each branch is factorized, shown in Fig. 8 (c).

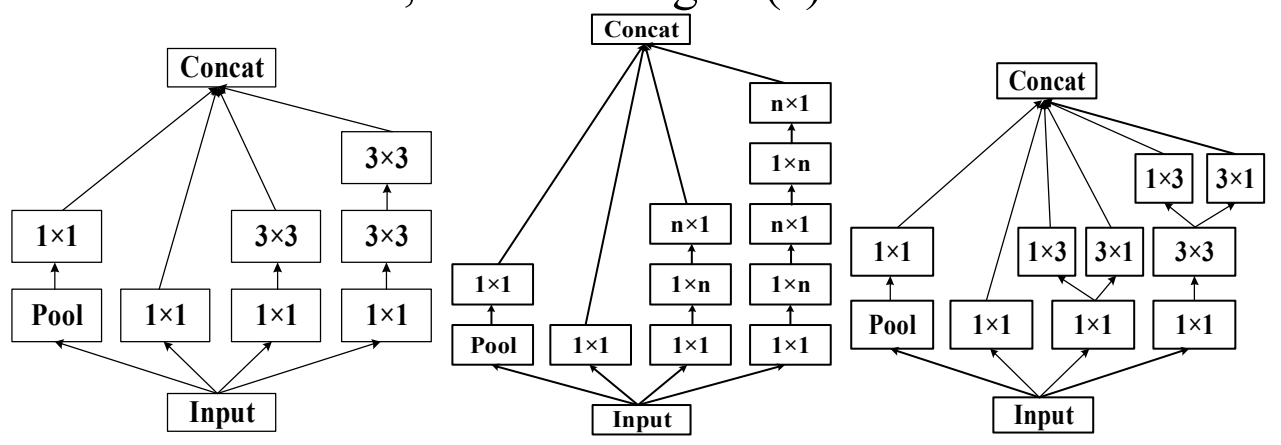

Fig. 8. Inception v2 module. (a) Each 5 x 5 convolution is replaced by two 3 x 3 convolutions. (b) n x n convolution is replaced by a 1 x n convolution and a n x 1 convolution. (c) Inception modules with the last convolutional layer is factorized.

#### *3) Inception v3*

Inception v3 [35] integrates major innovations mentioned in Inception v2. And factorizing 5 × 5 and 3 × 3 convolution kernels into two one-dimensional ones (one by seven and seven by one, one by three and three by one, respectively). This operation accelerates the training and further increases the depth of networks and the non-linearity of networks. Besides, the input size of the network changed from 224 by 224 to 299 by 299. And Inception v3 utilizes RMSProp as the optimizer.

#### *4) Inception v4 and Inception-ResNet*

Inception v4 modules [36] are based upon that of Inception v3. The architecture of Inception v4 is more concise and utilizes more Inception modules. Experimental evaluation proved that Inception v4 is better than its predecessors.

In addition, ResNet structure [37] is harnessed to extend the depth of Inception networks, namely Inception-ResNet-v1 and Inception-ResNet-v2. Experiments proved that they could improve the training speed and performance.

### *E. ResNet*

Theoretically, Deep Neural Networks (DNN) outperform shallow ones as the former can extract more complicated and sufficient features of images. However, with the increase of layers, DNNs are prone to cause gradient vanishing, gradient exploding problems, etc. He *et al.* [37] proposed a 34-layer Residual Network in 2016, which is the winner of the ILSVRC 2015 image classification and object detection algorithm. The performance of ResNet exceeds the GoogLeNet Inception v3.

One of the significant contributions of ResNet is the two-layer residual block constructed by the shortcut connection, as shown in Fig. 9 (a) below.

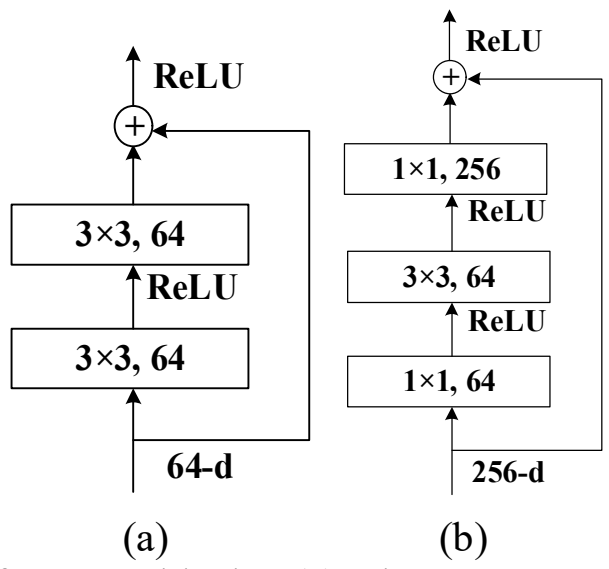

Fig. 9. Structure of ResNet blocks. (a) The structure of two-layer residual block. (b) The structure of three-layer residual block

50-layer ResNet, 101-layer ResNet, and 152-layer ResNet utilize three-layer residual blocks, as shown in the Fig. 9 (b) above, instead of two-layer one. Three-layer residual block is also called the bottleneck module because the two ends of the block are narrower than the middle. Using 1 × 1 convolution kernel can not only reduce the number of parameters in the network but also greatly improve the network's nonlinearity.

A lot of experiments in [37] have proved that ResNet can mitigate the gradient vanishing problem without degeneration in deep neural networks since the gradient can directly flow through shortcut connections.

Based upon ResNet, many studies have managed to improve the performance of the original ResNet, such as pre-activation ResNet [38], wide ResNet [39], stochastic depth ResNets (SDR) [40], and ResNet in ResNet (RiR) [41].

### *F. DCGAN*

Generative Adversarial Network (GAN) [42] is an unsupervised model proposed by Goodfellow *et al.* in 2014. GAN contains a generative model G and a discriminative model D. The model G with random noise z generates a sample G(z) that subjects to the data distribution $P_{data}$ learned by G. The model D can determine whether the input sample is real data x or generated data G(z). Both G and D can be nonlinear functions, such as deep neural networks. The aim of G is to generate data as real as possible; nevertheless, the aim of D is to distinguish the fake data generated by G from the real data. There exists an interestingly adversarial relationship between the generative network and the discriminative network. This idea originates from game theory, in which the two sides use their strategies to achieve the goal of winning. The procedure is shown in Fig. 10.

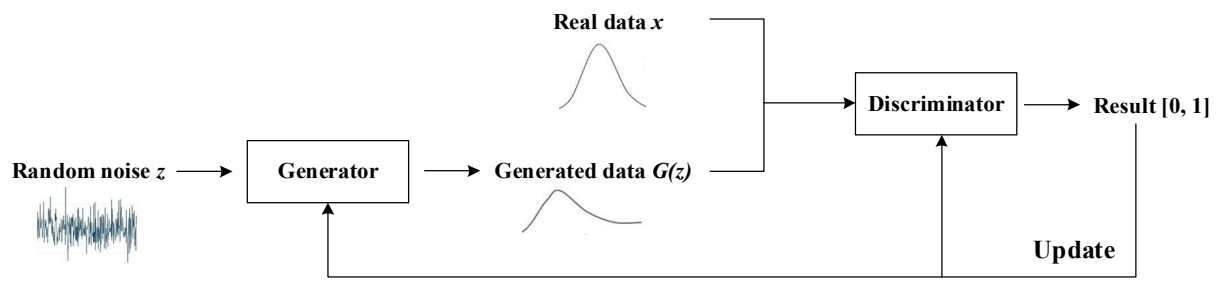

Fig. 10. The flowchart of GAN

Radford *et al.* [43] proposed Deep Convolutional Generative Adversarial Network (DCGAN) in 2015. The generator of DCGAN on Large-scale Scene Understanding (LSUN) dataset is implemented by using deep convolutional neural networks, the structure of which is shown in the figure below.

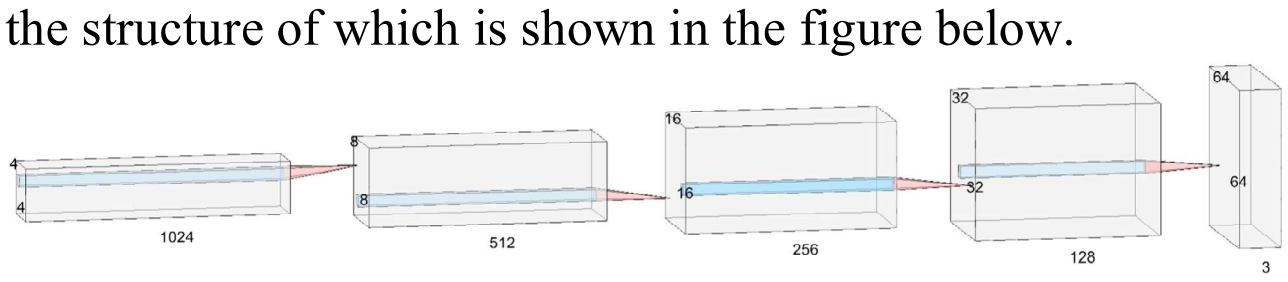

Fig. 11. DCGAN generator used for LSUN scene modeling

In Fig. 11, the generative model of DCGAN performs up-sampling by "fractionally-strided convolution". As shown in Fig. 12 (a), supposing that there is a 3 × 3 input, and the size of the output is expected to be larger than 3 × 3, then the 3 × 3 input can be expanded by inserting zero between pixels. After expanding to a 5 × 5 size, performing convolution, shown in Fig. 12 (b), can obtain an output larger than 3 × 3.

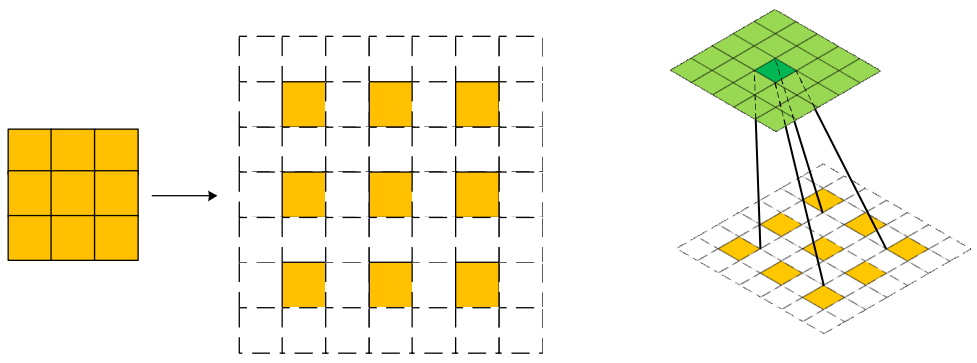

(a) (b)

Fig. 12. An example of fractionally-strided convolution. (a) Inserting zero between 3 × 3 kernel points. (b) Convolving the 7 × 7 graph

### *G. MobileNets*

MobileNets are a series of lightweight models proposed by Google for embedded devices such as mobile phones. They use depth-wise separable convolutions and several advanced techniques to build thin deep neural networks. There are three versions of MobileNets to date, namely MobileNet v1 [44], MobileNet v2 [45], and MobileNet v3 [46].

#### *1) MobileNet v1*

MobileNet v1 [44] utilizes depth-wise separable convolutions proposed in Xception [26], which decomposes the standard convolution into depth-wise convolution and pointwise convolution (1 × 1 convolution), as shown in Fig. 13. Specifically, standard convolution applies each convolution kernel to all the channels of input. In contrast, depth-wise convolution applies each convolution kernel to only one channel of input, and then 1 × 1 convolution is used to combine the output of depth-wise convolution. This decomposition can substantially reduce the number of parameters.

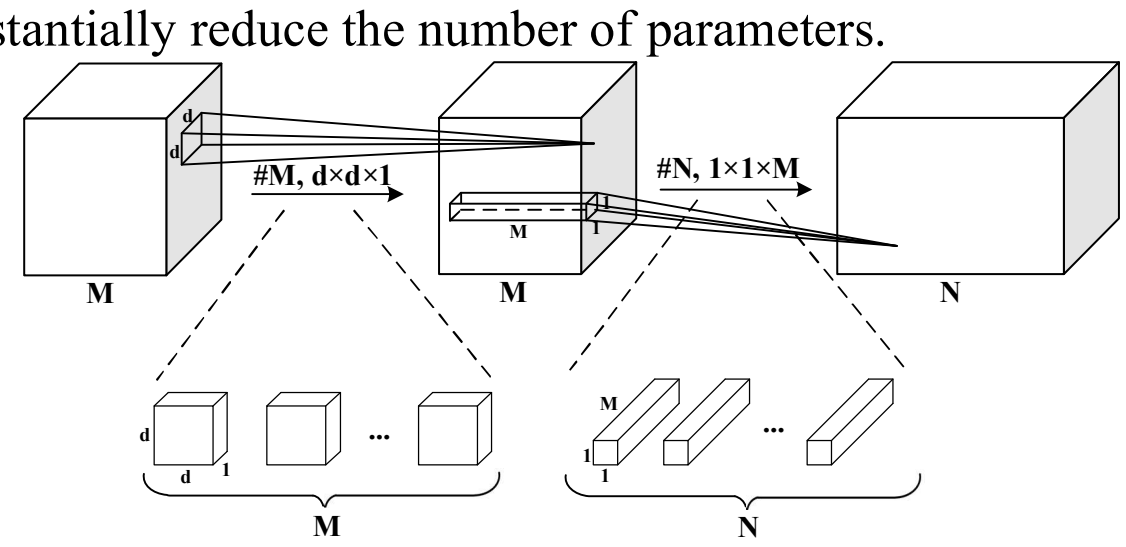

Fig. 13. Depth-wise separable convolutions in MobileNet v1. #M and #N represent the number of kernels of depth-wise convolution and pointwise convolution, respectively

MobileNet v1 also introduces the width multiplier to reduce the number of channels of each layer and the resolution multiplier to lower the resolution of the input image (feature map).

#### *2) MobileNet v2*

Based upon MobileNet v1, MobileNet v2 [45] mainly introduces two improvements: inverted residual blocks and linear bottleneck modules.

In Section 3.5, we have explained three-layer residual blocks, the purpose of which is to make use of 1 × 1 convolution to reduce the number of parameters involved in 3 × 3 convolution. In a word, the whole process of a residual block is channel compression—standard convolution—channel expansion. In MobileNet v2, an inverted residual block (seen in Fig. 14 (b)) is opposite to a residual block (seen in Fig. 14 (a)). The input of an inverted residual block is firstly convoluted by 1 × 1 convolution kernels for channel expansion, then convoluted by 3 × 3 depth-wise separable convolution, and finally convoluted by 1 × 1 convolution kernels to compress the number of channels back. Briefly speaking, the whole process of an inverted residual block is channel expansion—depth-wise separable convolution—channel compression. Also, due to the fact that depth-wise separable convolution cannot change the number of channels, which causes the number of input channels limits the feature extraction, inverted residual blocks are harnessed to handle the problem.

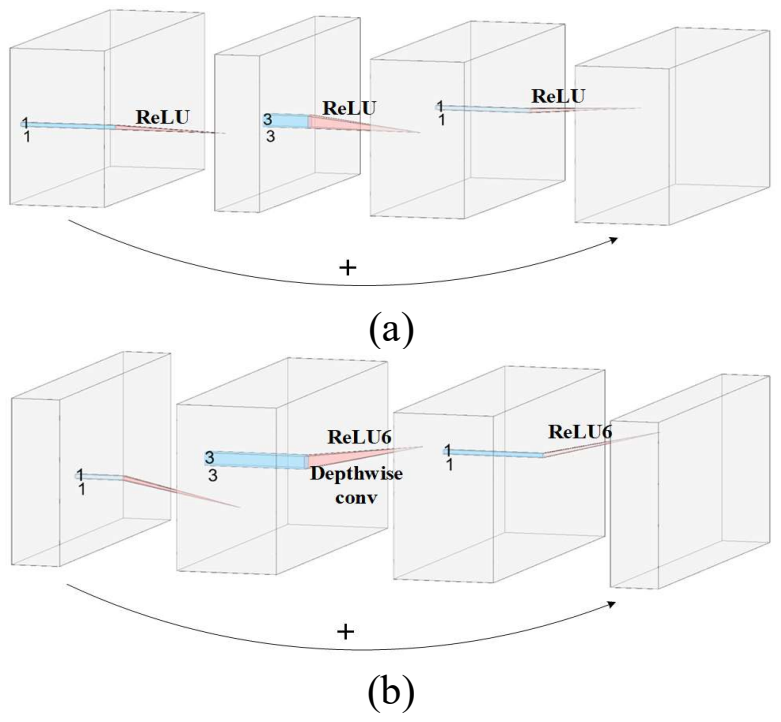

Fig. 14. (a) Residual block (b) MobileNet v2 block: inverted residual block

When performing the steps: channel expansion—depth-wise separable convolution—channel compression, one problem will be encountered after "channel compression". That is, it is easy to destroy the information when ReLU activation function is utilized in low-dimensional space, whereas it will not happen in high-dimensional space. Therefore, ReLU activation function following the second 1 × 1 convolution of inverted residual blocks is removed, and a linear transformation is adopted. Hence, this architecture is called the linear bottleneck module.

*3) MobileNet v3*

MobileNet v3 [46] achieves three improvements: network search combining platform-aware neural architecture search (platform-aware NAS) and NetAdapt algorithm [47], lightweight attention model based upon squeeze and excitation, and h-swish activation function.

For MobileNet v3, researchers use platform-aware NAS for block-wise search. Platform-aware NAS utilizes an RNN-based controller and hierarchical search space to find the structure of the global network. And then, the NetAdapt algorithm, complementary to platform-aware NAS, is used for layer-wise search. It can fine-tune to find the optimal number of filters in each layer.

MobileNet v3 makes use of the squeeze and excitation (SE) [48] to reweight the channels of each layer to achieve a lightweight attention model. As shown in Fig. 15, after the depth-wise convolution of an inverted residual block, the SE module is added. Global-pool operation is firstly performed, then following a fully-connected layer, the number of channels is reduced to 1/4. The second fully-connected layer is utilized to recover the number of channels and get the weight of each layer. Finally, multiply the weight and the depth-wise convolution to get a reweighted feature map. Howard *et al.* [46] proved that this operation could improve the accuracy without extra time cost.

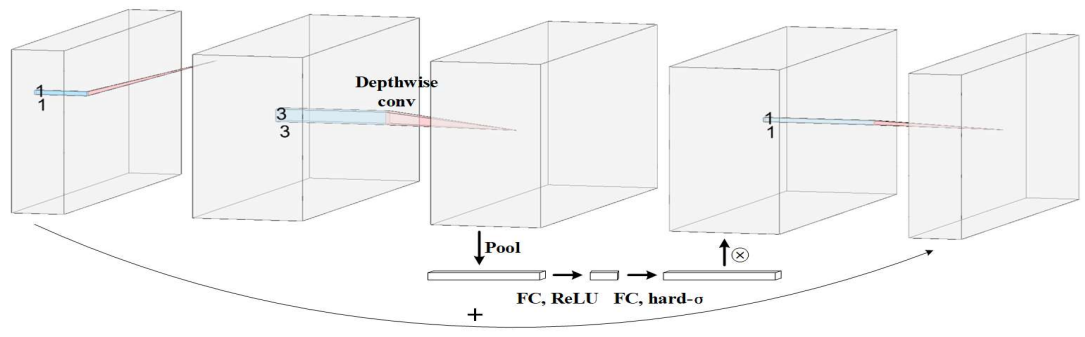

Fig. 15. MobileNet v3 block: MobileNet v2 block + Squeeze-and-Excite in the residual layer

The authors of MobileNet v3 figure out that swish activation function can improve the accuracy of the network compared with ReLU. Nevertheless, swish function costs too much computation, and hence, they put forward a hard version of swish (h-swish) to reduce computation with little loss of accuracy. However, they found that the benefits gained by h-swish only in the deep layer, and therefore, h-swish is only utilized in the second half of the model. Besides, they found that sigmoid can also be replaced by hard version of sigmoid (h-sigmoid).

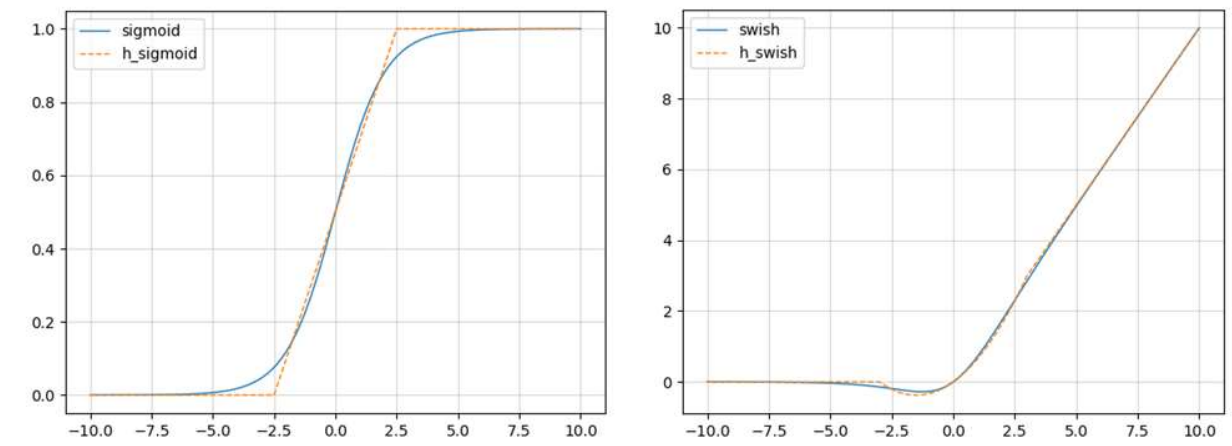

Fig. 16. The diagrams of sigmoid, h-sigmoid, swish and h-swish

## *H. ShuffleNets*

ShuffleNets are a series of CNN-based models proposed by MEGVII to solve the problem of insufficient computing power of mobile devices. These models combine pointwise group convolution, channel shuffle, and some other techniques, which significantly reduce the computational cost with little loss of accuracy. So far, there are two versions of ShuffleNets, namely ShuffleNet v1 [49] and ShuffleNet v2 [50].

*1) ShuffleNet v1*

ShuffleNet v1 [49] was proposed to construct a high-efficient CNN structure for resource-limited devices. There are two innovations: pointwise group convolution and channel shuffle.

The authors of ShuffleNet v1 reckon that Xception [26] and ResNeXt [29] are less efficient in extremely small networks since 1 × 1 convolution requires a lot of computing resources. Therefore, pointwise group convolution is proposed to reduce the computation complexity of 1 × 1 convolutions. Pointwise group convolution, shown in Fig. 17 (a), requires each convolution operation is only on the corresponding input channel group, which can reduce the computational complexity.

However, one problem is that pointwise group convolutions prevent feature maps between different groups from communicating with each other, which is harmful to extract representative feature maps. Therefore, channel shuffle operation, shown in Fig. 17 (b), is proposed to help the information in different groups flow to other groups randomly.

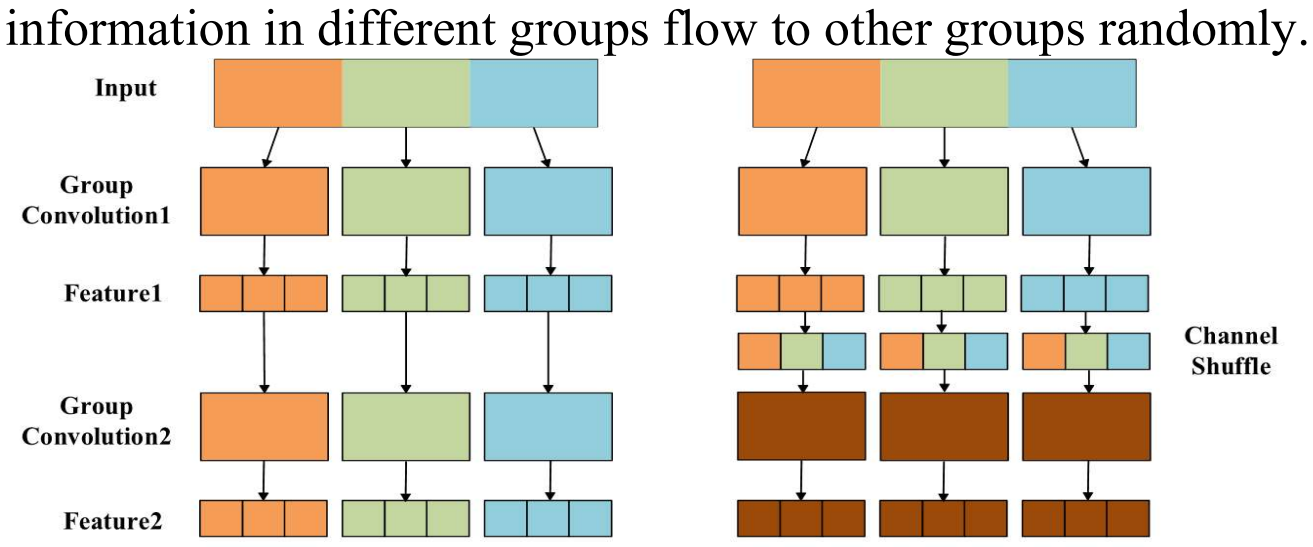

Fig. 17. Pointwise group convolution and channel shuffle in ShuffleNet v1 (a) Pointwise group convolution where different colors represent different groups. (b) After acquiring feature 1, channel shuffle operation is inserted to promote information exchanges between groups

Furthermore, ShuffleNet unit is proposed on the basis of channel shuffle operation. As shown in figure below, Fig. 18 (a) is a naïve residual block with depth-wise convolution (DWConv); Fig. 18 (b) replaces standard convolution with computational economical pointwise group convolution (GConv) and channel shuffle. Besides, the second ReLU activation function is removed; In Fig. 18 (c), a 3 × 3 average pooling with two strides is utilized in shortcut paths, and element-wise addition is replaced by concatenation. These tricks further reduce the number of parameters. Fig. 18 (c) is the architecture of the final ShuffleNet unit.

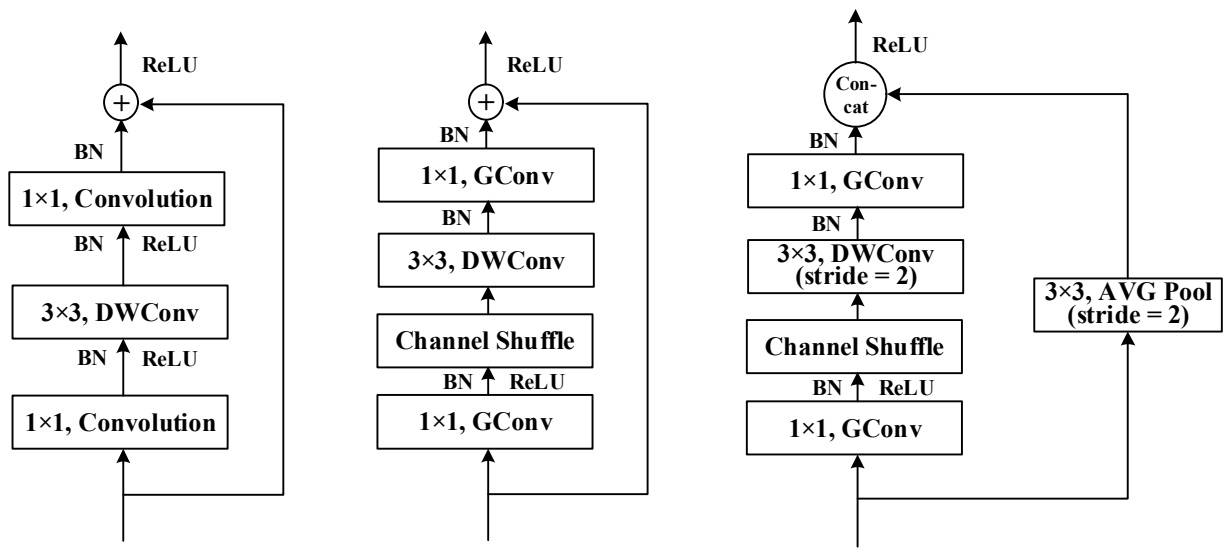

(a) (b) (c)

Fig. 18. ShuffleNet v1 units. (a) Naïve residual block with DWConv (b) ShuffleNet v1 unit with GConv and channel shuffle (c) ShuffleNet v1 unit with 3 x 3 average pooling and stride = 2.

*2) ShuffleNet v2*

The authors of ShuffleNet v2 [50] suggest that a lot of networks are dominated by the metric of computation complexity, i.e., FLOPs. Nevertheless, FLOPs should not be regarded as the only norm of evaluating the speed of networks because Memory Access Cost (MAC) is another crucial factor. They provide some guidelines for designing networks through experiments and uses these guidelines to build ShuffleNet v2. Four points are proposed to guide the design of networks:

a) Through experimental, compared with different input and output channel ratios of ShuffleNet v1 [49] and MobileNet v2 [45] on GPU and ARM platforms with 1 × 1 convolutional layers, it is found that the MAC is minimal when the number of input channels is equal to the number of output channels.

b) Changing the number of groups of convolution has an impact on network training speed. As the number of groups increases, the MAC increases, and the training speed decreases.

c) By adjusting fragmented structures of the networks and the number of convolutional layers in each basic structure, it is found that network fragmentation reduces the degree of parallelism, such as GoogLeNet series. Although multiple fragmented structures are able to improve accuracy, they can reduce efficiency on parallel computing powers like GPUs.

d) Elementwise operations (such as ReLU, tensor addition, offset addition, separation convolution, etc.) are non-negligible, i.e., they will consume a lot of time. Consequently, when designing networks, researchers should reduce the use of elementwise operations as much as possible.

ShuffleNet v2 unit is designed based upon the above four guidelines. Additionally, channel split is proposed in ShuffleNet v2. For each ShuffleNet v2 unit, the channels are firstly split into two branches, namely A and B. The details of the following process can be seen in Fig. 19 (a). For spatial downsampling, the unit is slightly modified, and the details can be seen in Fig. 19 (b).

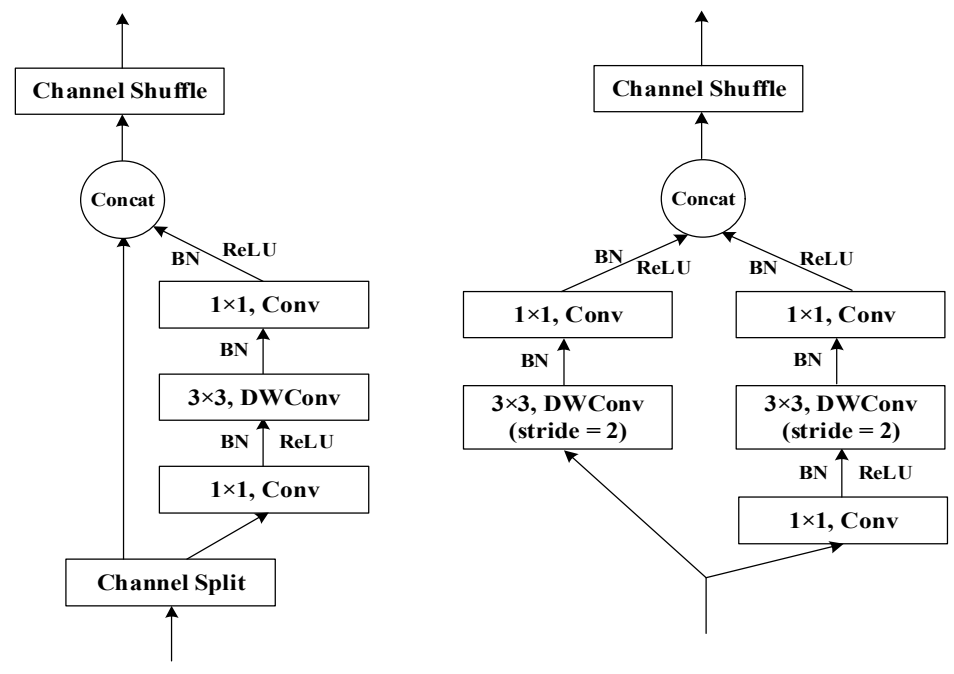

(a) (b)

Fig. 19. ShuffleNet v2 units (a) ShuffleNet v2 basic unit (b) ShuffleNet v2 unit for spatial down sampling

In summary, LeNet-5 is the first modern CNN utilized in handwritten digits recognition successfully. Its convolution, activation, pooling, and full connection have been widely used. AlexNet brought up deeper structures than LeNet-5, proposing some tricks like dropout and data augmentation, and harnessing ReLU activation function. VGGNet further proved that deeper networks usually work better, and provided a guideline for designing networks. GoogLeNet series proposed wider networks can also work. And large size convolution kernels can be replaced with small ones. Besides, factorizing convolution and ResNet are employed as well. ResNet makes extreme deep networks possible, which is able to mitigate gradient vanishing. DCGAN combines CNN with GAN, expanding the practical scenarios of both. Depth-wise separable convolution, inverted residual blocks, SENet-based lightweight attention model, platform-aware NAS, and NetAdapt algorithm are harnessed by MobileNets designing for mobile devices with limited computing power. ShuffleNet series is also invented for mobile devices, combining pointwise group convolution and channel shuffle. Moreover, ShuffleNet proved that except for FLOPs, MAC is another factor that can affect the speed of networks.

## *I. GhostNet*

As large amounts of redundant features are extracted by existing CNNs for image cognition, Han *et al.* [51] proposed GhostNet to reduce computational cost effectively. They found that there are many similar feature maps in traditional convolution layers. These feature maps are called ghost. Therefore, they leverage the cost-efficient GhostNet to reach state-of-the-art results. Two major contributions are as follows:

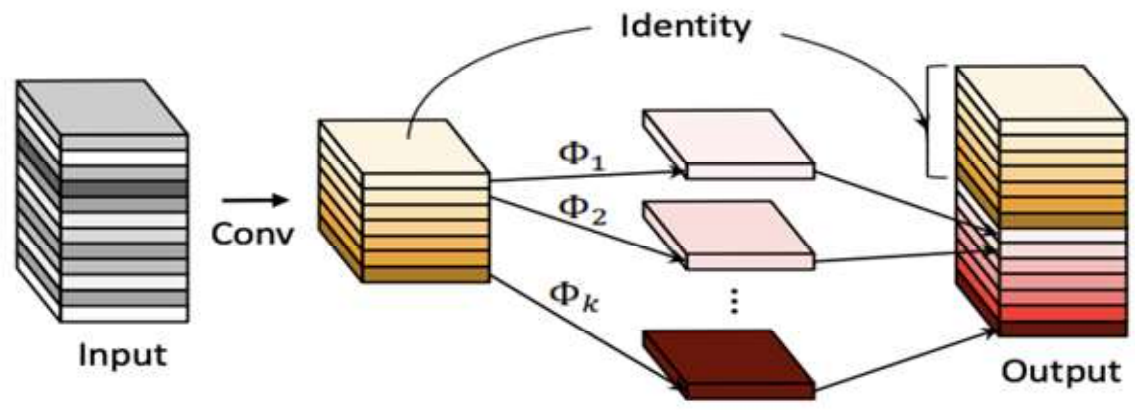

Fig. 20. The Ghost module. Where Φ represents linear transformation function [51]

They divide the traditional convolution layers into two parts. In the first part, less convolution kernels are directly used in feature extraction, which is the same as the original convolution. Then, these features are processed in linear transformation to acquire multiple feature maps. They proved that Ghost module applies to other CNN models.

## IV. Discussion and Experimental Analysis

### *A. Activation function*

#### *1) Discussion of Activation Function*

Convolutional neural networks can harness different activation functions to express complex features. Similar to the function of the neuron model of the human brain, the activation function here is a unit that determines which information should be transmitted to the next neuron. Each neuron in the neural network accepts the output value of the neurons from the previous layer as input, and passes the processed value to the next layer. In a multilayer neural network, there is a function between two layers. This function is called activation function, whose structure is shown in the following Fig. 21.

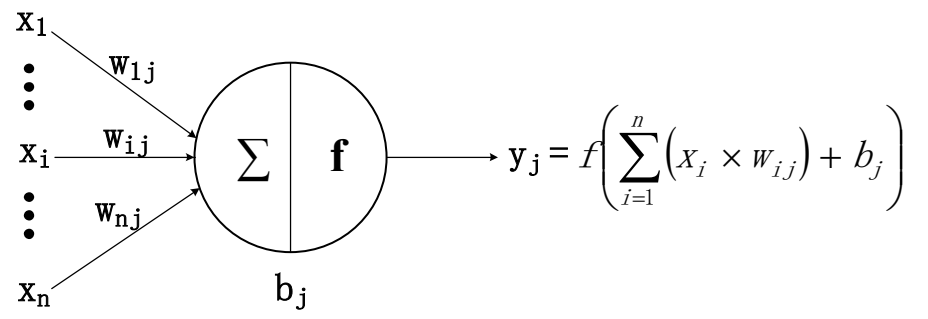

Fig. 21. General activation function structure

In this figure, $x_i$ represents the input feature; n features are input to the neuron j at the same time; $w_{ij}$ represents the weight value of the connection between the input feature $x_i$ and the neuron j; $b_j$ represents the internal state of the neuron j, which is the bias value; and $y_j$ is the output of the neuron j. $f(\cdot)$ is the activation function, which can be sigmoid function, tanh (x) function [10], Rectified Linear Unit [52], etc. [53]

If an activation function is not used or a linear function is used, the input of each layer will be a linear function of the output of the previous layer. In this case, He *et al.* [38] verify no matter how many layers the neural network has, the output is always a linear combination of the input, which means hidden layers have no effect. This situation is the primitive perceptron [2], [3], which has the limited learning ability. For this reason, the nonlinear functions are introduced as activation functions. Theoretically, the deep neural networks with nonlinear activation function can approximate any function, which greatly enhances the ability of neural networks to fit data.

In this section, we mainly focus on several frequently-used activation functions. To begin with, sigmoid function is one of the most typical non-linear activation functions with an overall S-shape (see Fig. 22 (a)). With x value approaching 0, the gradient becomes steeper. Sigmoid function can map a real number to (0, 1), so it can be used for binary classification problems. In addition, SENet [48] and MobileNet v3 [46] need to transform the output value to (0, 1) for attention mechanism, in which sigmoid is a good way to implement.

Different from sigmoid, tanh function [10] (see Fig. 22 (b)) can map a real number to (-1, 1). Since the mean value of the output of tanh is 0, it can achieve a kind of normalization. This makes the next layer easier to learn.

In addition, Rectified Linear Unit (ReLU) [52] (see Fig. 22 (c)) is another effective activation function. When x is less than 0, its function value is 0; when x is greater than or equal to 0, its function value is x itself. Compared to sigmoid function and tanh function, a significant advantage of using ReLU function is that it can speed up learning. Sigmoid and tanh involve in exponential operation that require division while computing derivatives, whereas the derivative of ReLU is a constant. Moreover, in the sigmoid and tanh function, if the value of x is too large or too small, the gradient of the function is pretty small, which can cause the function to converge slowly. However, when x is less than 0, the derivative of ReLU is 0, and when x is greater than 0, the derivative is 1, so it can obtain an ideal convergence effect. AlexNet [11], the best model in ILSVRC-2012, uses ReLU as the activation function of CNN-based model, which mitigates gradient vanishing problem when the network is deep, and verifies that the use of ReLU surpasses sigmoid in deep networks.

From what discussed above, we can find that ReLU does not consider the upper limit. In practice, we can set an upper limit, such as ReLU6 [54].

However, when x is less than 0, the gradient of ReLU is 0, which means the back-propagated error will be multiplied by 0, resulting in no error being passed to the previous layer. In this scenario, the neurons will be regarded as inactivated or dead. Therefore, some improved versions are proposed. Leaky ReLU (see Fig. 22 (d)) can reduce neuron inactivation. When x is less than 0, the output of Leaky ReLU is $x/a$, instead of zero, where 'a' is a fixed parameter in range $(1, +\infty)$.

Another variant of ReLU is PReLU [38] (see Fig. 22 (e)). Unlike Leaky ReLU, the slope of the negative part of PReLU is based upon the data, not a predefined one. He et *al.* [38] reckon that PReLU is the key to surpassing the level of human classification on the ImageNet 2012 classification dataset.

Exponential Linear Units (ELU) function [55] (see Fig. 22 (f)) is another improved version of ReLU. Since ReLU is non-negatively activated, the average value of its output is greater than 0. This problem will cause the offset of the next layer unit. ELU function has a negative value, so the average value of its output is close to 0, making the rate of convergence faster than ReLU. However, the negative part is a curve, which demands lots of complicated derivatives.

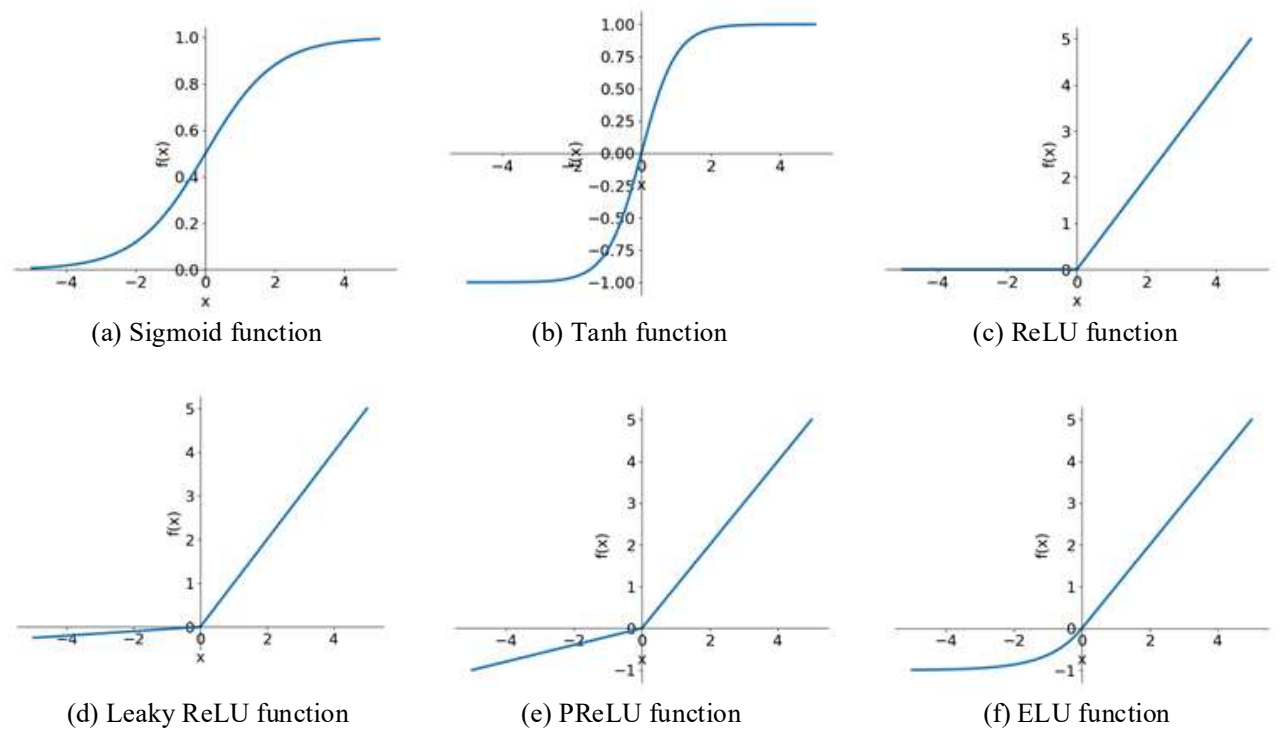

Fig. 22. Diagrams of Sigmoid, Tanh, ReLU, Leaky ReLU, PReLU, and ELU

#### *2) Experimental Evaluation*

To compare aforementioned activation functions, two classic CNN models, LeNet-5 [10] and VGG-16 [32], are tested on four benchmark datasets, including MNIST [10], Fashion-MNIST [56], CIFAR-10 [57] and CIFAR-100 [57]. LeNet-5 is the first modern but relatively shadow CNN model. In the following experiments, we train LeNet-5 from scratch. VGG-16 is a deeper, larger, and frequently-used model. We conduct our experiments on the basis of a pre-trained VGG-16 model without the last three layers on ImageNet [58].

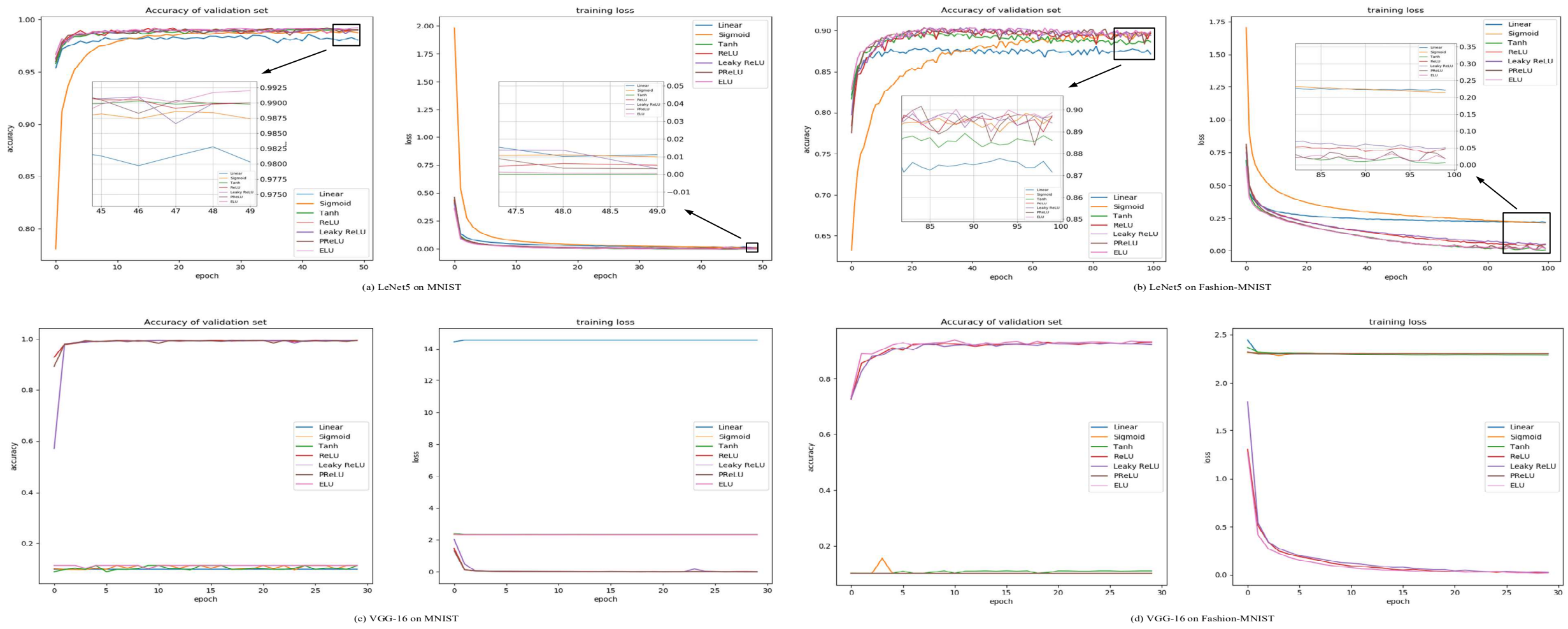

Fig. 23. The experimental results on seven activation functions, respectively. (a) The accuracy of validation set and training loss on MNIST trained by LeNet-5. (b) The accuracy of validation set and training loss on Fashion-MNIST trained by LeNet-5. (c) The accuracy of validation set and training loss on MNIST trained by VGG-16. (d) The accuracy of validation set and training loss on Fashion-MNIST trained by VGG-16.

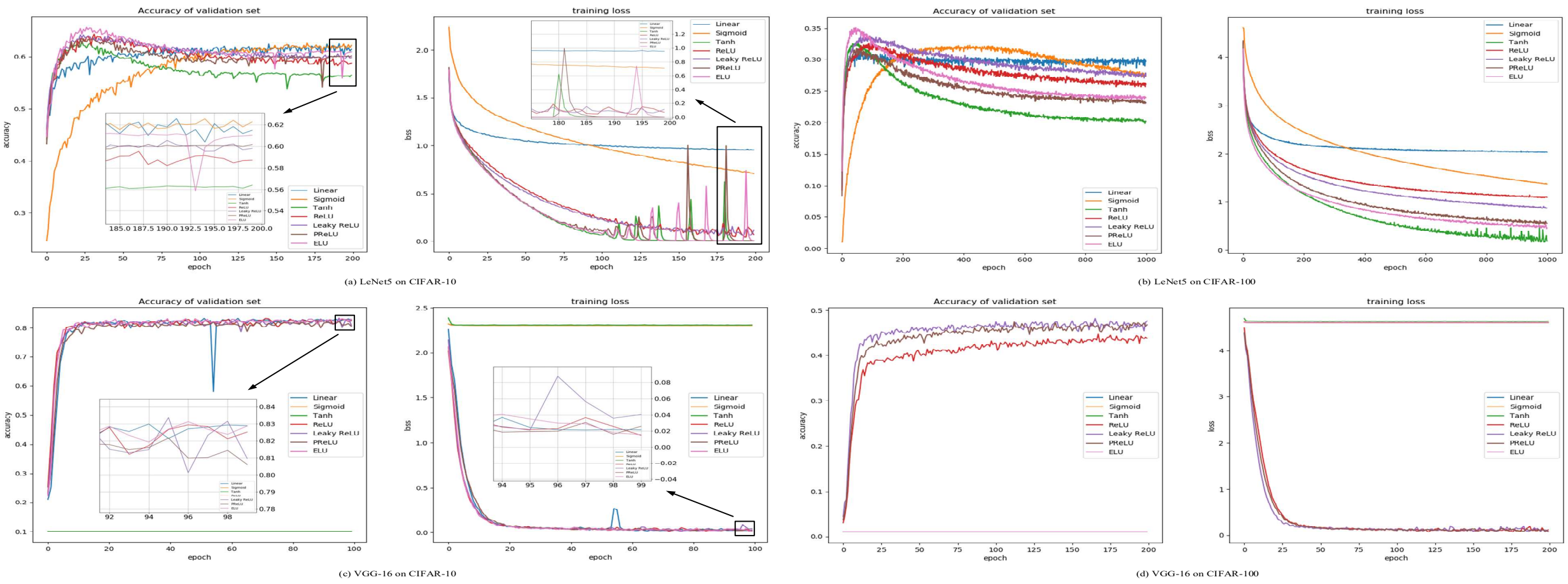

Fig. 24. The experimental results on seven activation functions, respectively. (a) The accuracy of validation set and training loss on CIFAR-10 trained by LeNet-5. (b) The accuracy of validation set and training loss CIFAR-100 trained by LeNet-5. (c) The accuracy of validation set and training loss on CIFAR-10 trained by VGG-16. (d) The accuracy of validation set and training loss on CIFAR-100 trained by VGG-16.

Both LeNet-5 and VGG-16 deploy softmax at the last layer for multi-classification. All experiments are tested on Intel Xeon E5-2640 v4 (X2), NVIDIA TITAN RTX (24GB), Python 3.5.6, and Keras 2.2.4.

*a) MNIST & Fashion-MNIST:* MNIST is a dataset of handwritten digits consisting of 10 categories, which has a training set of 60,000 examples and a test set of 10,000 examples. Each example is a 28 × 28 grayscale image, associated with a label from 10 classes, from 0 to 9. Fashion-MNIST dataset is a more complicated version of original MNIST, sharing the same image size, structure, and split. These two datasets are trained on LeNet-5 and VGG-16, and results are exhibited in Table I and Fig. 23. From the results, we can draw some meaningful conclusions.

- Linear activation function indeed lead to the worst performance. Therefore, when building a deep neural network (more than one layer), we need to add a non-linear function. If not, multiple layers, theoretically, are equal to one layer.
- Among these activation functions, the convergence speed of sigmoid is slowest. Usually, the final performance of sigmoid is not all that excellent. As a result, if we expect a fast convergence, sigmoid is not the best solution.
- From the perspective of accuracy, ELU possesses the best accuracy, but only a little better than ReLU, Leaky ReLU, and PReLU. In terms of training time, from Table I, ELU is prone to consume more time than ReLU and Leaky ReLU.
- ReLU and Leaky ReLU have better stability during training than PReLU and ELU.

*b) CIFAR-10 & CIFAR-100:* CIFAR-10 and CIFAR-100 are labeled subsets of the 80 million tiny images dataset [57], which are more complex than MNIST as well as Fashion-MNIST. CIFAR-10 dataset consists of 60000 32 × 32 RGB images in 10 classes, with 6000 images per class. The whole dataset is divided into 50000 training images and 10000 test images, i.e., each class has 5000 training images and 1000 test images. CIFAR-100 is like the CIFAR-10, except it has 100 classes containing 600 images per class. And each class has 500 training images and 100 test images. Similarly, we evaluate LeNet-5 and VGG-16 with different activation functions on these two datasets. The results can be seen in Table I and Fig.

TABLE I
COMPARATIVE RESULTS OF DIFFERENT ACTIVATION FUNCTIONS

| Model | Dataset | Data Augmentation | Loss | Optimizer | Batch size | epochs | Activation function | Validation set accuracy | Training time (s) |
|---|---|---|---|---|---|---|---|---|---|
| LeNet-5 | MNIST | — | Cross entropy | Adam | 256 | 50 | Linear | 98.56% | 92.89 |
| | | | | | | | Sigmoid | 98.94% | 95.51 |
| | | | | | | | Tanh | 99.03% | 92.92 |
| | | | | | | | ReLU | 99.18% | 95.04 |
| | | | | | | | Leaky ReLU | 99.10% | 99.82 |
| | | | | | | | PReLU | **99.20%** | 113.42 |
| | | | | | | | ELU | **99.20%** | 103.84 |
| | Fashion-MNIST | — | Cross entropy | Adam | 256 | 100 | Linear | 88.10% | 169.95 |
| | | | | | | | Sigmoid | 89.84% | 174.26 |
| | | | | | | | Tanh | 89.83% | 181.99 |
| | | | | | | | ReLU | 90.17% | 191.77 |
| | | | | | | | Leaky ReLU | 90.36% | 190.02 |
| | | | | | | | PReLU | 90.36% | 217.20 |
| | | | | | | | ELU | **90.37%** | 204.64 |
| | CIFAR-10 | — | Cross entropy | Adam | 256 | 200 | Linear | 62.56% | 614.89 |
| | | | | | | | Sigmoid | 62.65% | 569.25 |
| | | | | | | | Tanh | 62.94% | 575.07 |
| | | | | | | | ReLU | 64.40% | 550.35 |
| | | | | | | | Leaky ReLU | 64.27% | 582.08 |
| | | | | | | | PReLU | 63.61% | 650.75 |
| | | | | | | | ELU | **65.70%** | 626.51 |
| | CIFAR-100 | — | Cross entropy | Adam | 512 | 1000 | Linear | 31.24% | 2381.76 |
| | | | | | | | Sigmoid | 32.32% | 2355.64 |
| | | | | | | | Tanh | 32.69% | 2376.35 |
| | | | | | | | ReLU | 32.69% | 2418.18 |
| | | | | | | | Leaky ReLU | 33.81% | 2443.72 |
| | | | | | | | PReLU | 31.84% | 2615.02 |
| | | | | | | | ELU | **35.10%** | 2475.30 |
| VGG-16 (pre-trained) | MNIST | — | Cross entropy | Adam | 512 | 30 | Linear | 9.82% | 598.14 |
| | | | | | | | Sigmoid | 11.35% | 600.27 |
| | | | | | | | Tanh | 11.35% | 596.32 |
| | | | | | | | ReLU | **99.55%** | 606.83 |
| | | | | | | | Leaky ReLU | 99.48% | 608.91 |
| | | | | | | | PReLU | 99.45% | 607.27 |
| | | | | | | | ELU | 11.35% | 614.81 |
| | Fashion-MNIST | — | Cross entropy | Adam | 512 | 30 | Linear | 10.00% | 599.18 |
| | | | | | | | Sigmoid | 15.66% | 595.11 |
| | | | | | | | Tanh | 11.01% | 596.48 |
| | | | | | | | ReLU | 93.16% | 608.19 |
| | | | | | | | Leaky ReLU | 92.81% | 610.82 |
| | | | | | | | PReLU | 10.00% | 612.75 |
| | | | | | | | ELU | **93.87%** | 613.32 |
| | CIFAR-10 | — | Cross entropy | Adam | 512 | 100 | Linear | 83.25% | 958.74 |
| | | | | | | | Sigmoid | 10.00% | 957.62 |
| | | | | | | | Tanh | 10.00% | 957.97 |
| | | | | | | | ReLU | 83.22% | 957.74 |
| | | | | | | | Leaky ReLU | **83.37%** | 958.39 |
| | | | | | | | PReLU | 82.17% | 963.67 |
| | | | | | | | ELU | 83.14% | 968.60 |
| | CIFAR-100 | — | Cross entropy | Adam | 512 | 200 | Linear | 1.00% | 1897.14 |
| | | | | | | | Sigmoid | 1.00% | 1868.02 |
| | | | | | | | Tanh | 1.00% | 1897.76 |
| | | | | | | | ReLU | 44.77% | 1901.56 |
| | | | | | | | Leaky ReLU | **48.22%** | 1916.38 |
| | | | | | | | PReLU | 47.46% | 1922.29 |
| | | | | | | | ELU | 1.00% | 1917.75 |

24, from which we can get some conclusions.

- Tanh, PReLU, and ELU activation functions are more likely to bring about oscillation at the end of the training.
- When training a deep CNN model with pre-trained weights, it is hard to converge by the use of sigmoid and tanh activation functions.
- The models trained by Leaky ReLU and ELU have better accuracy than the others in the experiments. But sometimes, ELU may make networks learn nothing. More often than not, Leaky ReLU has better performance in terms of accuracy and training speed.

*3) Rules of Thumb for Selection*

- For binary classification problems, the last layer can harness sigmoid; for multi-classification problems, the last layer can harness softmax.
- Sigmoid and tanh functions sometimes should be avoided

because of the gradient vanishment. Usually, in hidden layers, ReLU or Leaky ReLU is a good choice.

- If you have no idea about choosing activation functions, feel free to try ReLU or Leaky ReLU.
- If a lot of neurons are inactivated in the training process, please try to utilize Leaky ReLU, PReLU, etc.
- The negative slope in Leaky ReLU can be set to 0.02 to speed up training.

### B. *Loss function*

Loss function or cost function is harnessed to calculate the distance between the predicted value and the actual value. Loss function is usually used as a learning criterion of the optimization problem. Loss function can be used with convolutional neural networks to deal with regression problems and classification problems, the goal of which is to minimize loss function. Common loss functions include Mean Absolute Error (MAE), Mean Square Error (MSE), Cross Entropy, etc.

#### 1) *Loss Function for Regression*

In convolutional neural networks, when dealing with regression problems, we are likely to use MAE or MSE.

MAE calculates the mean of the absolute error between the predicted value and the actual value; MSE calculates the mean of square error between them.

MAE is more robust to outliers than MSE, because MSE would calculate the square error of outliers. However, the result of MSE is derivable so that it can control the rate of update. The result of MAE is non-derivable, the update speed of which cannot be determined during optimization.

Therefore, if there are lots of outliers in the training set and they may have a negative impact on models, MAE is better than MSE. Otherwise, MSE should be considered.

#### 2) *Loss Function for Classification*

In convolutional neural networks, when it comes to classification tasks, there are many loss functions to handle.

The most typical one, called cross entropy loss, is used to evaluate the difference between the probability distribution obtained from the current training and the actual distribution. This function compares the predicted probability with the actual output value (0 or 1) in each class and calculate the penalty value based upon the distance from them. The penalty is logarithmic, so the function provides a smaller score (0.1 or 0.2) for smaller differences and a bigger score (0.9 or 1.0) for larger differences.

Cross entropy loss is also called softmax loss, which indicates it is always used in CNNs with a softmax layer. For example, AlexNet [11], Inception v1 [33], and ResNet [37] uses cross-entropy loss as the loss function in their original paper, which helped them reach state-of-the-art results.

However, cross entropy loss has some flaws. Cross entropy loss only cares about the correctness of the classification, not the degree of compactness within the same class or the margin between different classes. Hence, many loss functions are proposed to solve this problem.

Contrastive loss [59] enlarges the distance between different categories and minimizes the distance within the same categories. It can be used in dimensionality reduction in convolutional neural networks. After dimensionality reduction, the two samples that are originally similar are still similar in the feature space, whereas the two samples that are originally different are still different. Additionally, contrastive loss is widely used with convolutional neural networks in face recognition. It was firstly used in SiameseNet [60], and later was deployed in DeepID2 [61], DeepID2+ [62] and DeepID3 [63].

After contrastive loss, triplet loss was proposed by Schroff *et al.* in FaceNet [64], with which the CNN model can learn better face embeddings. The definition of the triplet loss function is based upon three images. These three images are anchor image, positive image, and negative image. The positive image and the anchor image are from the same person, whereas the negative image and the anchor image are from different people. Minimizing triplet loss is to make the distance between the anchor and the positive one closer, and make the distance between the anchor and the negative one further. Triplet loss is usually used with convolutional neural networks for fine-grained classification at the individual level, which requires model have ability to distinguish different individuals from the same category. Convolutional neural networks with triplet loss or its variants can be used in identification problems, such as face identification [65], [66], [64], person re-identification [67], [68], and vehicle re-identification [69].

Another one is center loss [70], which is an improvement based upon cross entropy. The purpose of center loss is to focus on the uniformity of the distribution within the same class. In order to make it evenly distributed around the center of the class, center loss adds an additional constraint to minimize the intra-class difference. Center loss was used with CNN in face recognition [70], image retrieval [71], person re-identification [72], speaker recognition [73], etc.

Another variant of cross entropy is large-margin softmax loss [74]. The purpose of it is also intra-class compression and inter-class separation. Large-margin softmax loss adds a margin between different classes, and introduces the margin regularity through the angle of the constraint weight matrix. Large-margin softmax loss was used in face recognition [74], emotion recognition [75], speaker verification [76], etc.

TABLE II
DIFFERENT LOSS FUNCTIONS FOR CONVOLUTIONAL NEURAL NETWORKS

| Loss | Brief Description |
|---|---|
| Mean absolute error | Calculate the mean of absolute error of samples. |
| Mean square error | Calculate the mean of square error of samples. |
| Cross entropy loss | Calculate the difference between the probability distribution and the actual distribution. [11] [33] [37] |
| Contrastive loss | Enlarge the distance between different categories and minimize the distance within the same categories. [60] [61] [62] [63] |
| Triplet loss | Minimize the distance between anchor samples and positive samples, and enlarge the distance between anchor samples and negative samples. [64] [65] [66] [67] [68] |
| Center loss | Minimize intra-class distance. [70] [71] [72] [73] |
| Large-margin softmax loss | Focus on intra-class compression and inter-class separation. [74] [75] [76] |

*3) Rules of Thumb for Selection*

- When using CNN models to deal with regression problems, we can choose L1 loss or L2 loss as the loss function.
- When dealing with classification problems, we can select the rest of the loss functions.
- Cross entropy loss is the most popular choice, usually appearing in CNN models with a softmax layer in the end.
- If the compactness within the class or the margin between different classes is concerned, the improvements based upon cross entropy loss can be considered, like center loss and large-margin softmax loss.
- The selection of loss function in CNNs also depends on the application scenario. For example, when it comes to face recognition, contrastive loss and triplet loss are turned out to be the commonly-used ones nowadays.

## *C. Optimizer*

In convolutional neural networks, we often need to optimize non-convex functions. Mathematical methods require huge computing power, so optimizers are used in the training process to minimize the loss function for getting optimal network parameters within acceptable time. Common optimization algorithms are Momentum, RMSprop, Adam, etc.

*1) Gradient Descent*

There are three kinds of gradient descent methods that we can use to train our CNN models: Batch Gradient Descent (BGD), Stochastic Gradient Descent (SGD), and Mini-Batch Gradient Descent (MBGD).

The BGD indicates a whole batch of data need to be calculated to get a gradient for each update, which can ensure convergence to the global optimum of the convex plane and the local optimum of the non-convex plane. However, it's pretty slow to use BGD because the average gradient of whole batch samples should be calculated. Also, it can be tricky for data that is not suitable for in-memory calculation. Hence, BGD is hardly utilized in training CNN-based models in practice.

On the contrary, SGD only use one sample for each update. It is apparent that the time of SGD for each update greatly less than BGD because only one sample's gradient is needed to calculate. In this case, SGD is suitable for online learning [77]. However, SGD is quickly updated with high variance, which will cause the objective function to oscillate severely. On the one hand, the oscillation of the calculation can make the gradient calculation jump out of the local optimum, and finally reach a better point; on the other hand, SGD may never converge because of endless oscillation.

Based on BGD and SGD, MBGD was proposed, which combines the advantages of BGD and SGD. MBGD uses a small batch of samples for each update, so that it can not only perform more efficient gradient decent than BGD, but also reduce the variance, making the convergence more stable.

Among these three methods, MBGD is the most popular one. Lots of classic CNN models use it to train their networks in original papers, like AlexNet [11], VGG [32], Inception v2 [34], ResNet [37] and DenseNet [78]. It has also been leveraged in FaceNet [64], DeepID [79], and DeepID2 [61].

*2) Gradient Descent Optimization Algorithms*

On the basis of MBGD, a series of effective algorithms for optimization are proposed to accelerate model training process. A proportion of them are presented as follows.

Qian *et al.* proposed the Momentum algorithm [80]. It simulates physical momentum, using the exponentially weighted average of the gradient to update weights. If the gradient in one dimension is much larger than the gradient in another dimension, the learning process will become unbalanced. The Momentum algorithm can prevent oscillations in one dimension, thereby achieving faster convergence. Some classic CNN models like VGG [32], Inception v1 [33], and Residual networks [37] use momentum in their original paper.

However, for the Momentum algorithm, blindly following gradient descent is a problem. Nesterov Accelerated Gradient (NAG) algorithm [81] gives the Momentum algorithm a predictability that makes it slow down before the slope becomes positive. By getting the approximate gradient of the next position, it can adjust the step size in advance. Nesterov Accelerated Gradient has been used to train CNN-based models in many tasks [82], [83], [84].

Adagrad algorithm [85] is another optimization algorithm based upon gradients. It can adapt the learning rate to parameters, performing smaller updates (i.e., a low learning rate) for frequent feature-related parameters, and performing larger-step updates (i.e., a high learning rate) for infrequent ones. Therefore, it is very suitable for processing sparse data. One of the main advantages of Adagrad is that there is no need to adjust the learning rate manually. In most cases, we just use 0.01 as the default learning rate [50]. FaceNet [64] uses Adagrad as the optimizer in training.

Adadelta algorithm [86] is an extension of the Adagrad, designed to reduce its monotonically decreasing learning rate. It does not merely accumulate all squared gradients but sets a fixed size window to limit the number of accumulated squared gradients. At the same time, the sum of gradients is recursively defined as the decaying average of all previous squared gradients, rather than directly storing the previous squared gradients. Adadelta are leveraged in many tasks [87], [88], [89].

Root Mean Square prop (RMSprop) algorithm [90] is also designed to solve the problem of the radically diminishing learning rate in the Adagrad algorithm. MobileNet [44], Inception v3 [35], and Inception v4 [36] achieved their best models using RMSprop.

Another frequently-used optimizer is Adaptive Moment Estimation (Adam) [91] It is essentially an algorithm formed by combining the Momentum and the RMSprop. Adam stores both the exponential decay average of the past square gradients like the Adadelta algorithm and the average exponential decay average of the past gradients like the Momentum algorithm. Practice has proved that Adam algorithm works well on many problems and is applicable to many different convolutional neural network structures [92], [93], [88].

AdaMax algorithm [91] is a variant of Adam that makes the boundary range of the learning rate simpler, and it has been used to train CNN models [94], [95].

TABLE III
COMPARATIVE RESULTS OF DIFFERENT OPTIMIZERS

| Model | Last three layers | Dataset | Data Augmentation | Loss | Activation function | Batch size | epochs | Optimizer | Validation set accuracy | Training time (s) |
|---|---|---|---|---|---|---|---|---|---|---|
| VGG-16 (pre-trained) | 512, 256, 10 | CIFAR-10 | — | Cross entropy | ReLU | 512 | 100 | MBGD | 85.70% | 926.24 |
| | | | | | | | | Momentum | 86.37% | 947.18 |
| | | | | | | | | Nesterov | 84.32% | 945.92 |
| | | | | | | | | Adagrad | 84.68% | 950.72 |
| | | | | | | | | Adadelta | 86.06% | 965.90 |
| | | | | | | | | RMSprop | **87.32%** | 959.33 |
| | | | | | | | | Adam | 83.09% | 953.46 |
| | | | | | | | | Adamax | 86.18% | 960.83 |
| | | | | | | | | Nadam | 86.26% | 968.72 |
| | | | | | | | | AMSgrad | 83.25% | 951.28 |

Nesterov-accelerated Adaptive Moment Estimation (Nadam) [96] is a combination of Adam and NAG. Nadam has a stronger constraint on the learning rate and a direct impact on the update of the gradient. Nadam is used in many tasks [97], [98], [99].

AMSGrad [100] is an improvement on Adagrad. The author of AMSGrad algorithm found that there were errors in the update rules of the Adam algorithm, which caused it to fail to converge to the optimal in some cases. Therefore, AMSGrad algorithm uses the maximum value of the past squared gradient instead of the original exponential average to update the parameters. AMSGrad has been used to train CNN in many tasks [101], [102], [103].

*3) Experimental Evaluation*

In the experiment, we tested ten kinds of optimizers—mini-batch gradient decent, Momentum, Nesterov, Adagrad, Adadelta, RMSprop, Adam, Adamax, Nadam, and AMSgrad on CIFAR-10 data set. The last nine optimizers are based upon mini batch. The format of the CIFAR -10 data set is the same as the experiment in the section 2.B. We also do our experiments on the basis of a pre-trained VGG-16 model without the last three layers on ImageNet [58]. The results can be seen in Table III and Fig. 25, from which we can get some conclusions.

- Almost all optimizers that we tested can make the CNN-based model converge at the end of the training.
- The convergence rate of mini-batch gradient decent is slowest, even if it can converge at the end.
- In the experiment, we find that Nesterov, Adagrad, RMSprop, Adamax, and Nadam oscillate and even cannot converge during training. In the further experiments (see Fig. 26.), we find that learning rate has huge impact on convergence.
- Nesterov, RMSprop, and Nadam are likely to create oscillation, but it is this characteristic that may help models jump out of local optima.

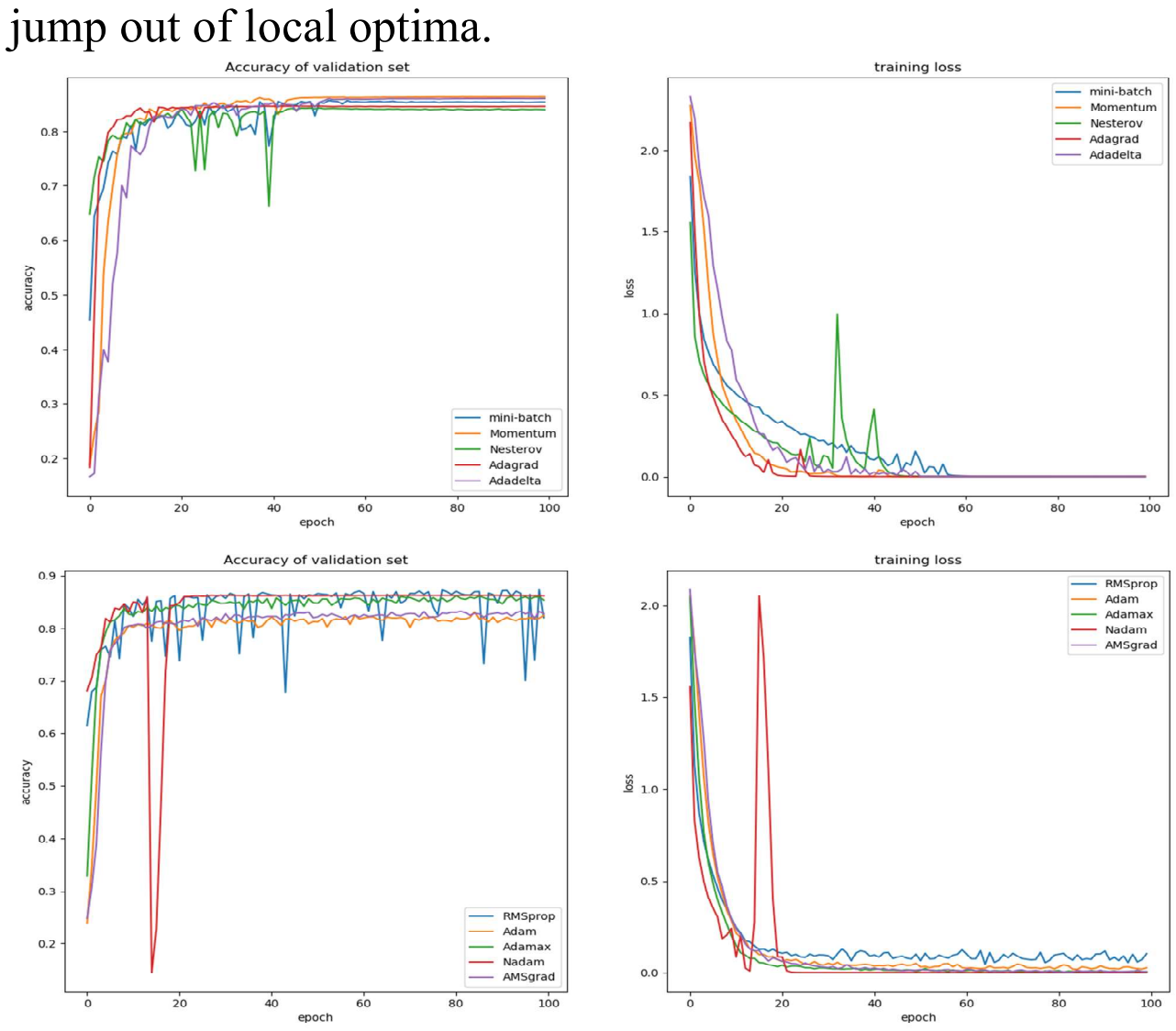

Fig. 25. The accuracy of validation set and training loss on CIFAR-10 trained by VGG-16 with ten different optimizers, respectively.

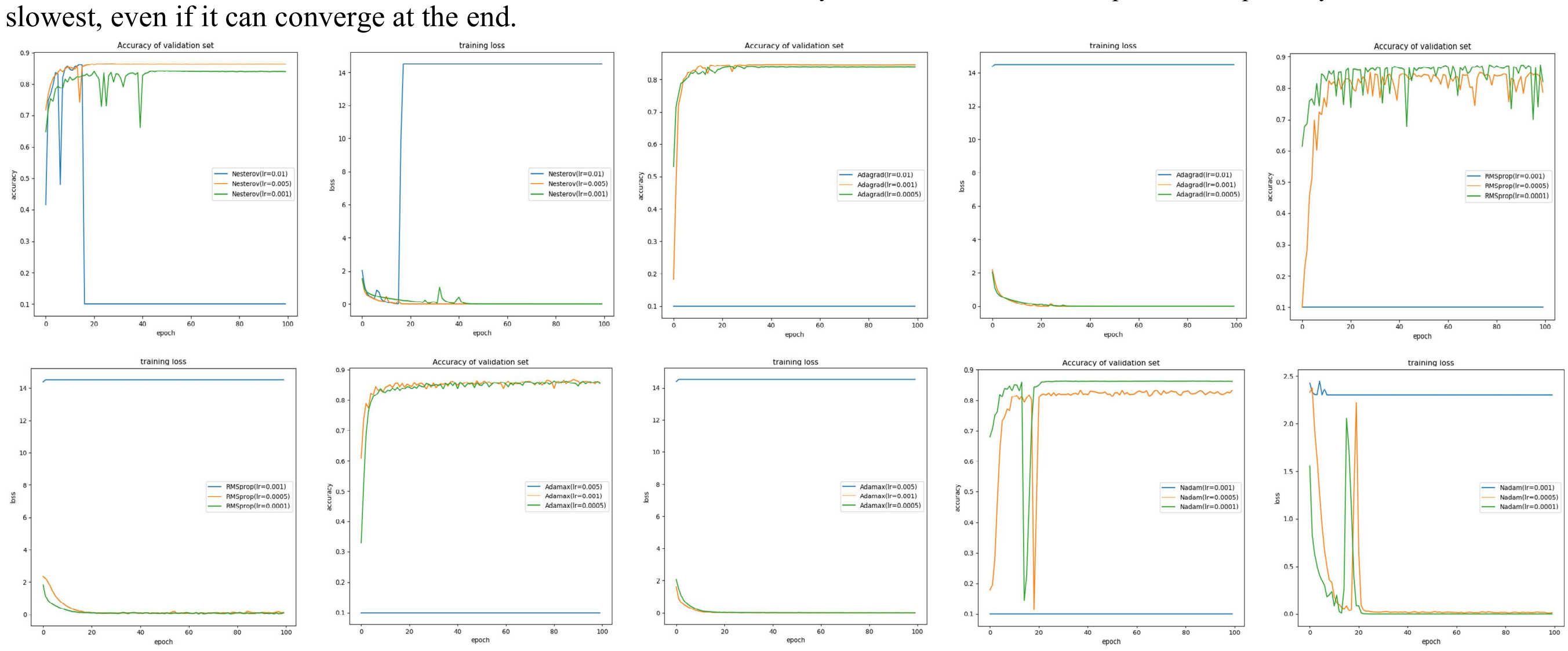

Fig. 26. The accuracy of validation set and training loss on CIFAR-10 trained by VGG-16 with Nesterov, Adagrad, RMSprop, Adamax, or Nadam optimizer with different learning rates, respectively.

#### 4) *Rules of Thumb for Selection*

- Mini batch should be used in order to make a trade-off between the computing cost and the accuracy of each update.
- The performance of optimizers is closely related to data distribution, so please feel free to try different optimizers mentioned above.
- If excessive oscillation or divergence occurs, lowering the learning rate may be a good choice.

## V. APPLICATIONS OF CNN

Convolutional neural network is one of the crucial concepts in the field of deep learning. In the era of big data, different from traditional approaches, CNN is able to harness a massive amount of data to achieve a promising result. Hence, there are lots of applications that come up. It can be used not only in the processing of two-dimensional images but also in one-dimensional and multi-dimensional scenarios.

### A. *Applications of one-dimensional CNN*

One-dimensional CNN (1D CNN) typically leverages one-dimensional convolutional kernels to process one-dimensional data. It is very effective when extracting the feature from a fixed-length segment in the whole dataset where the position of the feature does not matter. Therefore, 1D CNN can be applied to time series prediction and signal identification, for example.

#### 1) *Time Series Prediction*

1D CNN can be applied to time series prediction of data, such as electrocardiogram (ECG) time series, weather forecast, and traffic flow prediction. Erdenebayar *et al.* [104] proposed a method based on 1D-CNN to predict atrial fibrillation using short-term ECG data automatically. Harbola *et al.* [105] proposed 1D Single (1DS) CNN for predicting dominant wind speed and direction. Han *et al.* [106] applied 1D-CNN on short-term highway traffic flow prediction. 1D-CNN is used to extract spatial features of traffic flow, which is combined with temporal features to predict the traffic flow.

#### 2) *Signal Identification*

Signal identification is to discriminate the input signal according to the feature that CNN learned from training data. It can be applied to ECG signal identification, structural damage identification, and system fault identification. Zhang *et al.* [107] proposed a multi-resolution one-dimensional convolutional neural network structure to identify arrhythmia and other diseases based on ECG data. Abdeljaber *et al.* [108] proposed a direct damage identification method based on 1D CNN that can apply to the original environmental vibration signals. Abdeljaber *et al.* [109] designed a compact 1D CNN used in fault detection and severity identification of ball bearings.

### B. *Applications of two-dimensional CNN*

#### 1) *Image Classification*

Image classification is the task of classifying an image into a class category. CNN represents a breakthrough in this field.

LeNet-5 [10] is regarded as the first application used in hand-written digits classification. AlexNet [11] made CNN-based classification approaches get off the ground. Then, Simonyan [32] *et al.* emphasize the importance of depth, but these primitive CNNs are not more than ten layers. Afterward, deeper network structures emerged, such as GoogLeNet [33] and VGGNets [32], which significantly improve the accuracy in classification tasks.

In 2014, He *et al.* [110] proposed the SPP-Net that inserts a pyramid pooling layer between the last convolution layer and the fully-connected layer, making the size of different input images get the same size outputs. In 2015, He *et al.* [37] proposed ResNet to solve the degradation problems and made it possible to train deeper neural networks. In 2017, Chen *et al.* [111] proposed a Double Path Network (DPN) for image classification by analyzing the similarities and differences between ResNet [37] and DenseNet [112]. DPN not only shares the same image features but also ensures the flexibility of structure feature extraction by double path. In 2018, Facebook opened the source code of ResNeXt-101 [113] and extended the number of layers of ResNeXt to 101, which achieved state-of-the-art results on ImageNet.

Also, CNN can be deployed in medical image classification [114], [115], traffic scenes related classification[116], [117], etc. [118], [119] Li *et al.* [114] designed a custom CNN with shallow convolution layers to the classification of Interstitial Lung Disease (ILD). Jiang *et al.* [115] proposed a method based on SE-ResNet modules to classify breast cancer tissues. Bruno *et al.* [116] applied Inception networks to the classification of traffic signal signs. Madan *et al.* [117] proposed a different preprocessing method to classify traffic signals.

#### 2) *Object Detection*

Object detection is the task based on image classification. Systems not only need to identify which category the input image belongs to, but also need to mark it with a bounding box. The development process of object detection based on deep learning is shown in Fig. 27. The approaches of object detection can be divided into one-stage approaches, like YOLO [120],[121],[122], SSD [123], CornerNet [124],[125], and two-stage approaches, like R-CNN [126], Fast R-CNN [127], Faster R-CNN [128].

In the two-stage object detection, the region proposals are selected in advance, and then the objects are classified by CNN. In 2014, Girshick *et al.* [126] used region proposal and CNN to replace the sliding window and manual feature extraction used in traditional object detection and designed the R-CNN framework, which made a breakthrough in object detection. Then, Girshick *et al.* [127] summarizing the shortcomings of R-CNN [126] and drawing lessons from the SPP-Net [110], proposed Fast R-CNN, which introduced the Regions of Interest (ROI) Pooling layer, making the network faster. Besides, Fast R-CNN shares convolution features between object classification and bounding box regression. However, Fast R-CNN still retains the selective search algorithm of R-CNN's region proposals. In 2015, Ren *et al.* [128] proposed Faster R-CNN, which adds the selection of region proposals to make it faster. An essential contribution of Faster R-CNN is to introduce an RPN network at the end of the convolutional layer. In 2016, Lin *et al.* [129] added Feature Pyramid Network (FPN) to Faster R-CNN, where multi-scale features can be fused through the feature pyramid in the forward process.

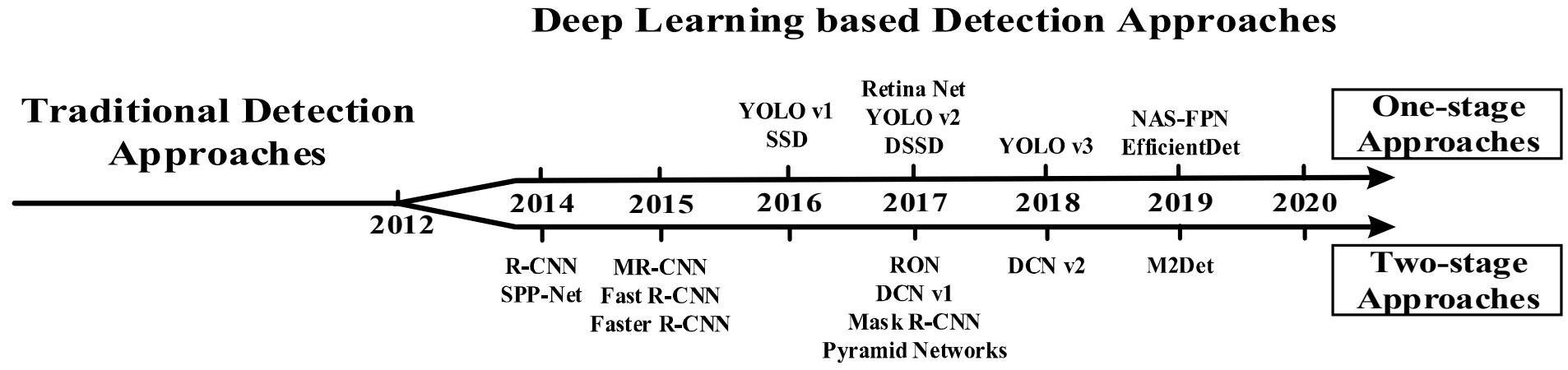

Fig. 27. Object detection milestones based on deep learning

In one stage, the model directly returns the category probability and position coordinates of the objects. Redmon *et al.* regarded object detection as a regression problem and proposed YOLO v1 [120], which directly utilizes a single neural network to predict bounding boxes and the category of objects. Afterward, YOLO v2 [121] proposed a new classification model darknet-19, which includes 19 convolutional layers and five max-pooling layers. Batch normalization layers are added after each convolution layer, which is beneficial to stable training and fast convergence. YOLO v3 [122] was proposed to remove the max-pooling layers and the fully-connected layers, using $1 \times 1$ and $3 \times 3$ convolution and shortcut connections. Besides, YOLO v3 borrows the idea from FPN to achieve multi-scale feature fusion. For the benefits of the structure of YOLO v3, many classic networks replace the backend of it to achieve better results. All of the aforementioned approaches leverage anchor boxes to determine where objects are. Their performance hinges on the choice of anchor boxes, and a large number of hyperparameters are introduced. Therefore, Law *et al.* [124] proposed CornerNet, which abandons anchor boxes and directly predicts the top-left corner and bottom-right corner of bounding boxes of objects. In order to decide which two corners in different categories are paired with each other, and an embedding vector is introduced. Then, CornerNet-Lite [125] optimized CornerNet in terms of detection speed and accuracy.

*3) Image Segmentation*

Image segmentation is the task that divides an image into different areas. It has to mark the boundaries of different semantic entities in an image. The image segmentation task completed by CNN is shown in Fig 28.

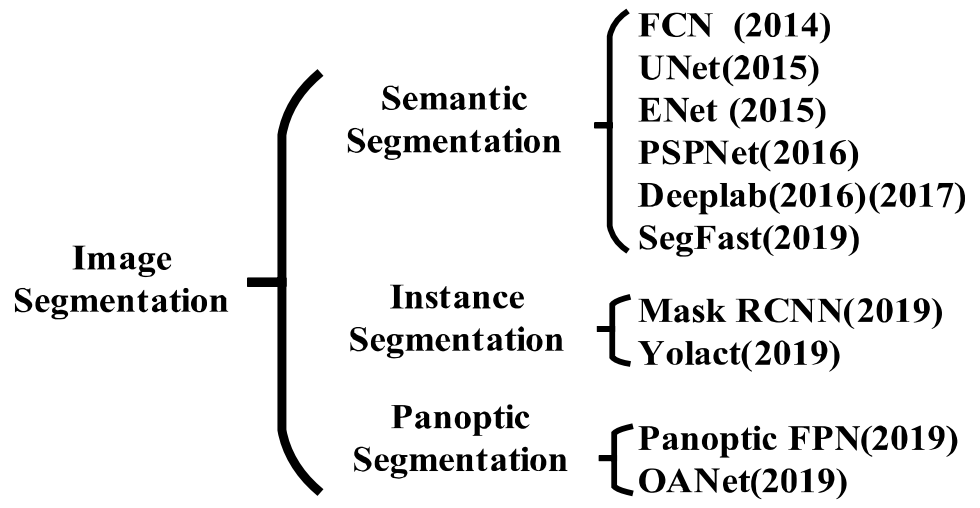

Fig. 28. Applications of CNN in image segmentation

In 2014, Long *et al.* [130] proposed the concept of Fully Convolutional Networks (FCN) and applied CNN structures to image semantic segmentation for the first time. In 2015, Ronneberger *et al.* [131] proposed U-Net, which has more multi-scale features and has been applied to medical image segmentation. Besides, ENet [132], PSPNet [133], etc. [134], [135] were proposed to handle specific problems.

In terms of instance segmentation tasks, He *et al.* [136] proposed Mask-RCNN that shares convolution features between two tasks through the cascade structure. In consideration of real time, Bolya *et al.* [137] based on RetinaNet [138] harnessed ResNet-101 and FPN to fuse multi-scale features.

For panoptic segmentation tasks, it was first proposed by Kirillov *et al.* [139] in 2018. They proposed panoramic FPN [140] in 2019, which combines FPN network with Mask-RCNN to generate a branch of semantic segmentation. In the same year, Liu *et al.* [141] proposed OANet that also introduced the FPN based on Mask-RCNN, but the difference is that they designed an end-to-end network.

*4) Face Recognition*

Face recognition is a biometric identification technique based on the features of the human face. The development history of deep face recognition is shown in Fig. 29. DeepFace [142] and DeepID [79] achieved excellent results on the LFW [74] data set, surpassing humans for the first time in the unconstrained scenarios. Henceforth, deep learning-based approaches received much more attention. The process of DeepFace proposed by Taigman et al. [142] is detection, alignment, extraction, and classification. After detecting the face, using three-dimensional alignment generate a $152 \times 152$ image as the input of CNN. Taigman *et al.* [142] leveraged Siamese network to train the model, which obtained state-of-the-art results. Unlike DeepFace, DeepID directly inputs two face images into CNN to extract feature vectors for classification. DeepID2 [61] introduces classification loss and verification loss. Based upon DeepID2 [61], DeepID2+ [62] adds the auxiliary loss between convolutional layers. DeepID3 [63] proposed two kinds of structures, which can be constructed by stacked convolutions of VGGNet or Inception modules.

The aforementioned approaches harness the standard softmax loss function. More recently, improvements in face recognition are basically focused on the loss function. FaceNet [85] proposed by Google in 2015 utilizes 22-layer CNN and 200 million pictures, including eight million people, to train a model. In order to learn more efficient embeddings, FaceNet replaces softmax with triplet loss. Besides, VGGFace [65] also deploys triplet loss to train the model. Besides, there are various loss functions harnessed to reach better results, like L-softmax loss, SphereFace, ArcFace, and large margin cosine loss, which can be seen in Fig. 29. [74], [143], [144], [145]

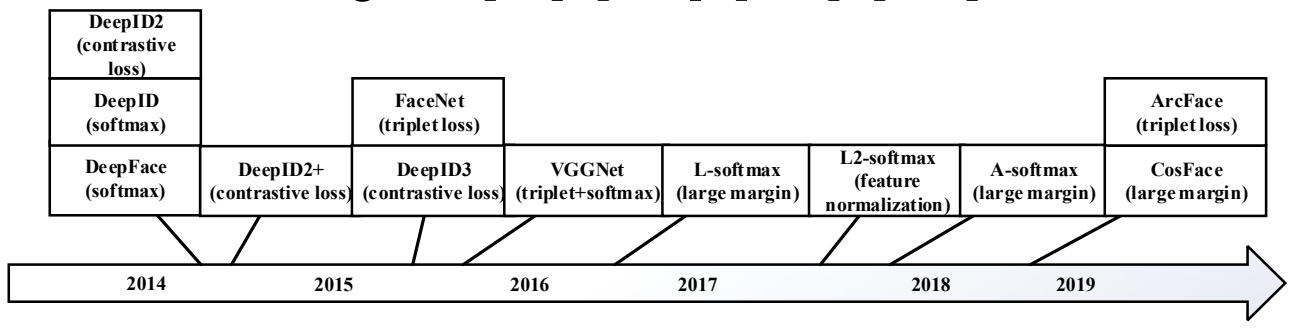

Fig. 29. The development history of deep face recognition

### *C. Applications of multi-dimensional CNN*

In theory, CNN can be leveraged in data with any dimension. However, since high dimensional data is hard to understand for human, multi-dimensional CNN is not too common in over three dimensions. Therefore, to better explain the key points, we take applications of three-dimensional CNN (3D CNN), for instance. It does not mean that higher dimensions are infeasible.

#### *1) Human Action Recognition*

Human Action recognition refers to the automatic recognition of human actions in videos by machines. As a kind of rich visual information, action recognition method based on human body posture is used in many jobs. Cao *et al.* [146] utilized 3D CNN to extract features of joints instead of the whole body, and then these features are fed into a linear SVM for classification. Another approach [147] is to extend 2D image convolutions to 3D video convolutions to extract spatio-temporal patterns. Huang *et al.* [148] designed a 3D-CNN structure to carry out sign language recognition that the 3D convolutional layers automatically extract distinct spatial and temporal features from the video stream, which are input to the fully connected layer for classification. In addition, it is also possible to integrate the features extracted from 3D CNN and 2D CNN. Huang *et al.* [149] fused the 3D convolutional pose stream with the 2D convolutional appearance stream that provides more discriminative human action information.

#### *2) Object Recognition/Detection*

In 2015, Wu *et al.* [150] proposed a generative 3D CNN of shape named 3D ShapeNet, which can be applied to object detection of RGBD images. In the same year, Maturana *et al.* [151]proposed VoxNet, which is a 3D CNN architecture that contains two 3D convolutional layers, pooling layer, and two fully-connected layers. In 2016, Song *et al.* [152] proposed a 3D region proposal network to learn the geometric feature of objects and designed a Joint Object Recognition Network that fuses the output of VGGNet and the 3D CNN to learn 3D bounding box and object classification jointly. In 2017, Zhou *et al.* [153] proposed a single end-to-end VoxelNet for point-cloud-based 3D detection. VoxelNet contains feature learning layers, convolution middle layers, and RPN. Each convolution middle layer uses 3D convolutions, batch normalization layers, and ReLU to aggregate voxel-wise features. Pastor *et al.* [154] designed TactNet3D, harnessing tactile features to recognize objects.

In addition, high dimensional images, like X-rays and CT images, can be detected by 3D CNN. A lot of practitioners [155], [156] are dedicated to these jobs.

## VI. PROSPECTS FOR CNN

With the rapid development of CNN, some methods are proposed to refine CNN, including model compression, security, network architecture search, etc. In addition, convolutional neural networks have many obvious disadvantages, like losing spatial information and other limitations. New structures need to be introduced to handle these problems. Based on the points mentioned above, this section briefly introduces several promising trends of CNN.

### *A. Model Compression*

Over the past decade, convolutional neural networks have been designed with various modules, which helped CNN reach good accuracy. However, high accuracy typically relies on extreme deep and wide architectures, which makes it challenging to deploy the models to embedded devices. Therefore, model compression is one of the possible ways to handle this problem, including low-rank approximation, network pruning, and network quantization.

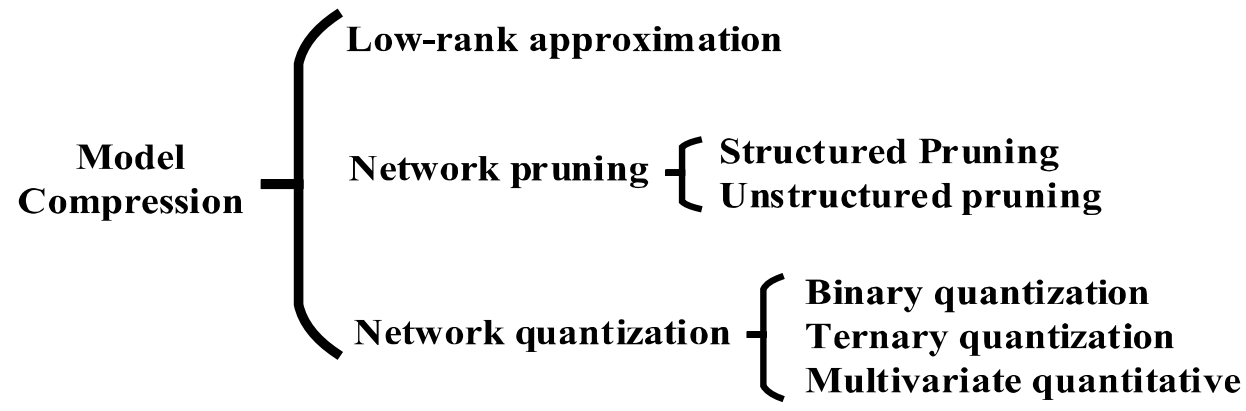

Fig. 30. Three directions of model compression

The methods of low-rank approximation consider the weight matrix of the original network as the full-rank matrix and then decomposes it into a low-rank matrix to approximate the effect of the full-rank matrix. Jaderberg *et al.* [157] proposed a structure-independent method using cross-channel redundancy or filter redundancy to reconstruct a low-rank filter. Sindhwani *et al.* [158] proposed a structural transformation network to learn a large class of structural parameter matrices characterized by low displacement rank through a unified framework.

Besides, according to the granularity of networks, pruning can be divided into structured pruning and unstructured pruning. Han *et al.* [159] proposed deep pruning for CNN to prune networks by learning essential connections. However, the fine-grain pruning method increases the irregular sparsity of convolution kernels and computational cost. Therefore, many pruning methods focus on coarse granularity. Mao *et al.* [160] deemed that coarse pruning reaches a better compression ratio than fine-grained pruning. Through weighing the relationship between sparsity and precision, they provide some advice on how to ensure accuracy when structural pruning.

Network quantization is another way to reduce computational cost, including binary quantization, ternary quantization, and multivariate quantization. In 2016, Rastegari *et al.* [161] proposed Binary-Weight Networks and XNOR-Networks. The former makes the values of networks approximate to binary values, and the latter approximate convolutions by binary operations. Lin *et al.* [162] use a linear combination of multiple binary weight bases to approximate full-precision weights and multiple binary activation to reduce information loss, which suppresses the prediction accuracy degradation caused by previous binary CNN. Zhu *et al.* [163] proposed that ternary quantization could alleviate the accuracy degradation. Besides, multivariate quantization is leveraged to represent weights by several values [164], [165].

As increasing networks use 1 × 1 convolution, low-rank approximation is difficult to achieve model compression. Network pruning is a major practical way in model compression tasks. Binary quantization tremendously reduces the model size with the cost of losing accuracy. Hence, ternary quantization and multivariate quantization are harnessed to strike a proper balance between model size and accuracy.

### *B. Security of CNN*

There are many applications of CNN in daily life, including security identification system [166],[167], medical image identification [168],[156],[155], traffic sign recognition [169], and license plate recognition [170]. These applications are highly related to life and property security. Once models are disturbed or destroyed, the consequences will be severe. Therefore, the security of CNN is expected to be attached great importance. More precisely, researchers [171],[172],[173],[174] have proposed some methods to deceive CNN, resulting in a sharp drop in the accuracy. These methods can be classified into two categories: data poisoning and adversarial attacks.

Data poisoning indicates that poisoning the training data during the training phase. Poison refers to the insertion of noise data into the training data. It is not easily distinguished at the image level and has no abnormalities found during the training process. Whereas in test stages, the trained model would reveal the problem of low accuracy. Furthermore, the noise data can even be fine-tuned so that the model can identify certain targets incorrectly. Liao *et al.* [172] introduced that generated perturbation masks are injected into the training samples as a backdoor to deceive the model. Backdoor injection does not interfere with normal behaviors but stimulates the backdoor instance to misclassify specific targets. Liu *et al.* [175] proposed a method of fine-tuning and pruning what can effectively defend against the threat of backdoor injection. The combination of pruning and fine-tuning succeeds in suppressing, even eliminating the effects of the backdoors.

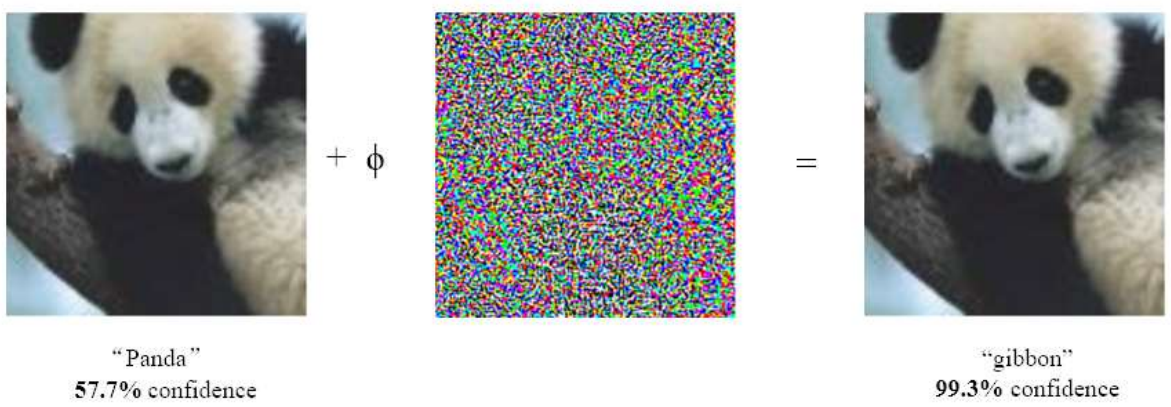

Fig. 31. A demonstration of fast adversarial example generation [174]

Adversarial attack is also one of the threats faced by deep neural networks. In Fig. 31, some noise is added to a normal image. Although naked eyes cannot distinguish the difference between two images, the CNN based model cannot recognize them as the same. Goodfellow et *al.* [174] reckoned that the main factor for the vulnerability of neural networks is linear characteristics, like ReLU, Maxout, etc. The cheap, analytical disturbance of these linear models would damage the neural network. Besides, they proposed a fast gradient notation to generate adversarial examples and found that many models misclassified these examples. Akhtar *et al.* [176] listed three directions of defense against adversarial attacks, which are respectively improved on training examples, modified trained networks, and additional networks. First, for the training examples, adversarial examples can be utilized to enhance the robustness of models. Second, network architectures can be adjusted to ignore noise. Last, additional networks can be used to help the backbone network against adversarial attacks.

### *C. Network Architecture Search*

Network Architecture Search (NAS) is another method to realize automatic machine learning of CNN. NAS constructs a search space through design choices, such as the number of kernels, skip connections, etc. Besides, NAS finds a suitable optimizer to control the search process in the search space. As shown in Fig. 32, NAS could be divided into NAS with agents and without agents. Due to the high demand for NAS on computing resources, the integrated models consist of learned optimal convolutional layers in the small-scale data sets. Small-scale data sets are the agents that generate the overall model, so this approach is the NAS with agents. The agentless NAS refers to learning the whole model directly on large-scale data sets.

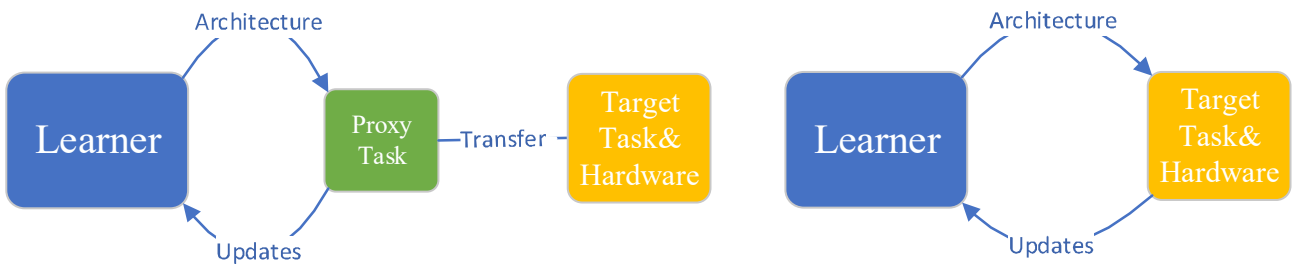

Fig. 32. The procedure of NAS. [179] (a) NAS with agents (b) NAS without agents.

In 2017, Google Inc. [177] proposed a machine learning search algorithm that uses reinforcement learning to maximize the target network and implements an auto-built network on CIFAR-10 data set, achieving similar precision and speed to networks with similar structure. Nevertheless, this approach is computationally expensive. Pham *et al.* [178] proposed Efficient Neural Architecture Search (ENAS), which shares parameters among sub-models and reduces resource requirements. Cai *et al.* [179] proposed ProxylessNAS, which is a path-level NAS method that has a model structure parameter layer at the end of the path and adds a binary gate before the output to reduce GPU utilization. It can directly learn architectures on the large-scale data set. Additionally, there are many ways to reduce the search space of reinforcement learning. Tan *et al.* [180] designed Mobile Neural Architecture Search (MNAS) to solve the CNN inferring delay problem. They introduced a decomposed hierarchical search space and performed the reinforcement learning structural search algorithm on this space. Ghiasi *et al.* [181] proposed NAS-FPN by applying NAS to feature pyramid structure search of object detection. They combined scalable search space with NAS to reduce search space. The scalable search space can cover all possible cross-scale connections and generate multi-scale feature representations.

### *D. Capsule Neural Network*

A lot of impressive applications of CNN are emerging in modern society, from simple cat-dog classifiers [182] to sophisticated autonomous vehicles [181], [183], [184]. However, is CNN a panacea? When does it not work?

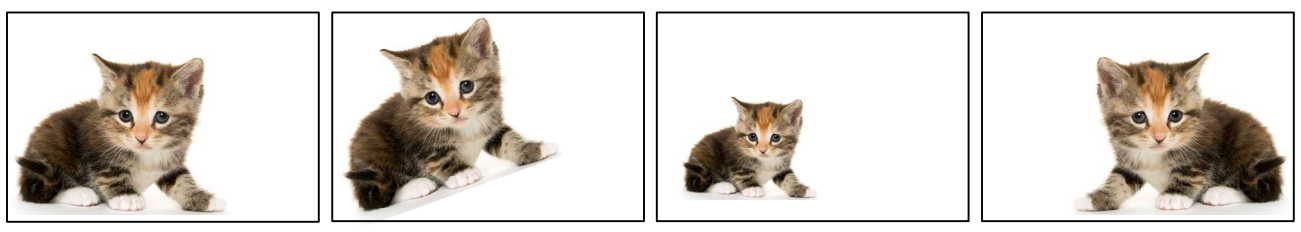

Fig. 33. The same cat in different ways

CNN is not sensitive to slight changes of images, such as rotation, scaling, and symmetry, which has been demonstrated by Azulay *et al.* [185]. However, when trained with Fig. 33 (a),

CNN cannot correctly recognize Fig. 33 (b), (c), or (d) is the same cat as the former, which is obvious to humans. This problem is caused by the architecture of CNN. Therefore, in order to teach a CNN system to recognize different patterns of one object, a massive amount of data should be fed, making up for the flaw of CNN architectures with diverse data. However, labeled data is typically hard to obtain. Although some tricks like data augmentation can bring about some effects, the improvement is relative limited.

Pooling layer is widely used in CNN for many advantages, but it ignores the relationship between the whole and the part. For effectively organizing network structures and solving the problem of spatial information loss of traditional CNN, Hinton *et al.* [186] proposed Capsule Neural Networks (CapsNet) where neurons on different layers focus on different entities or attributes, so that they add neurons to focus on the same category or attribute, similar to the structure of a capsule. When CapsNet is activated, the pathway between capsules forms tree structure composed of sparsely activated neurons. Each output of a capsule is a vector, the length of which represents the probability of the existence of an object. Therefore, the output features include the specific pose information of objects, which means that CapsNet has the ability to recognize the orientation. In addition, unsupervised CapsNet was created by Hinton *et al.* [187], called Stacked Capsule Autoencoder (SCAE). SCAE consists of four parts: Part Capsule Autoencoder (PCAE), Object Capsule Autoencoder (OCAE), and the decoders of PCAE and OCAE. PCAE is a CNN with a top-down attention mechanism. It can identify the pose and existence of capsules of different parts. OCAE is used to implement inference. SCAE can predict the activations of CapsNet directly based on the pose and the existence. Some experiments have proved that CapsNet is able to reach state-of-the-art results. Although it did not achieve satisfactory results on complicated large-scale data sets, like CIFAR-100 or ImageNet, we can see that it is potential.

## VII. Conclusion

Due to the advantages of convolutional neural networks, such as local connection, weight sharing, and down-sampling dimensionality reduction, they have been widely deployed in both research and industry projects. This paper provides a detailed survey on CNN, including common building blocks, classic networks, related functions, applications, and prospects.

First, we discuss basic building blocks of CNN and present how to construct a CNN-based model from scratch.

Second, some excellent networks are expounded. From them, we obtain some guidelines for devising novel networks from the perspective of accuracy and speed. More specifically, in terms of accuracy, deeper and wider neural structures are able to learn better representation than shallow ones. Besides, residual connections can be leveraged to build extremely deep neural networks, which can increase the ability to handle complex tasks. In terms of speed, dimension reduction and low-rank approximation are very handy tools.

Third, we introduce activation functions, loss functions, and optimizers for CNN. Through experimental analysis, several conclusions are reached. Also, we offer some rules of thumb for the selection of these functions.

Fourth, we discuss some typical applications of CNN. Different dimensional convolutions should be designed for various problems. Other than the most frequently-used two-dimensional CNN used for image-related tasks, one-dimensional and multi-dimensional CNN are harnessed in lots of scenarios as well.

Finally, even though convolutions possess many benefits and have been widely used, we reckon that it can be refined further in terms of model size, security, and easy hyperparameters selection. Moreover, there are lots of problems that convolution is hard to handle, such as low generalization ability, lack of equivariance, and poor crowded-scene results, so that several promising directions are pointed.